\documentclass[runningheads,a4paper]{llncs}

\usepackage{times}

\usepackage{graphicx}

\usepackage[scaled=1]{helvet} 
\usepackage{courier}

\usepackage{nameref}
\usepackage{amssymb}
\usepackage{graphicx}
\usepackage{textcomp}
\usepackage{tikz} 
\usepackage{amsmath}
\usepackage{comment}
\usepackage{enumitem}
\usepackage{tikz}
\usepackage[ruled,vlined]{algorithm2e}

\usepackage[T1,hyphens]{url}

\usepackage[draft,
            raiselinks=true,%
            bookmarks=false,%
            bookmarksopenlevel=1,%
            bookmarksopen=false,%
            bookmarksnumbered=true,%
            hyperindex=true,%
            plainpages=false,
            pdfpagelabels=true,
            pdfborder={0 0 0.5}]{hyperref} 
\pdfcompresslevel=9 \hypersetup{
 pdflang=en,
 colorlinks, 
 linkcolor=black,
 anchorcolor=black,
 filecolor=black,
 urlcolor=black,
 citecolor=black,
 bookmarksnumbered=true,
 pdfstartview=FitH
}
\usepackage{color}

\newcommand{\eod}{\ensuremath{\hfill\dashv}}
\newcommand{\fun}[1]{\ensuremath{\mbox{\sl #1}}}

\newcommand{\terms}[1]{\ensuremath{\fun{terms}(#1)}}
\newcommand{\vars}[1]{\ensuremath{\fun{vars}(#1)}}

\newenvironment{proofi} {\noindent\emph{Proof:}} {\hfill $\square$\vspace{0.2cm}}

\newcommand{\OB}{\ensuremath{\fun{$\mathbf{O}$}}}
\newcommand{\SO}{\ensuremath{\fun{$\mathbf{SO}$}}}
\newcommand{\RE}{\ensuremath{\fun{$\mathbf{R}$}}}
\newcommand{\EQ}{\ensuremath{\fun{$\mathbf{E}$}}}

\definecolor{gren}{rgb}{0.158, 0.488, 0.408}

\newcommand{\mlm}[1]{\textcolor{black}{#1}} 

\begin{document}

\title{On the $k$-Boundedness for Existential Rules}
\author{Stathis Delivorias \and Michel Lecl\`ere \and Marie-Laure Mugnier \and Federico Ulliana}
\institute{University of Montpellier, LIRMM,  CNRS, Inria\\
Montpellier, France }

\maketitle
\begin{abstract}  

The chase is a fundamental tool for existential rules. Several chase variants are known, which differ on how they handle redundancies possibly caused by the introduction of nulls. Given a chase variant, the halting problem takes as input a set of existential rules and asks if this set of rules ensures the termination of the chase for any factbase. It is well-known that this problem is undecidable for all known chase variants. The related problem of boundedness asks if a given set of existential rules is bounded, i.e., 
\mlm{whether there is a predefined upper bound on the number of (breadth-first) steps of the chase, independently from any factbase.}
 This problem is already undecidable in the specific case of datalog rules. However, knowing that a set of rules is bounded for some chase variant does not help much in practice if the bound is unknown. Hence, in this paper, we investigate the decidability of the $k$-boundedness problem, which asks whether a given set of rules is bounded by an integer $k$. 
 \mlm{We prove that $k$-boundedness is decidable for three chase variants, namely the oblivious, semi-oblivious and restricted chase. }

\end{abstract}

\mlm{\emph{This report is a revised version of the paper published at} RuleML+RR 2018.
}

\section{Introduction}


\sloppy
Existential rules (see  \cite{cgk08,blms09,cgl09} for the first papers and \cite{GottlobOPS12,MugnierT14} for introductory courses) are a positive fragment of first-order logic that generalizes the deductive database query language Datalog and knowledge representation formalisms such as Horn description logics (see e.g. \cite{dl-lite05,krh07,ltw09}).
 These rules offer the possibility to model the existence of unknown individuals by means of existentially quantified variables in rule heads, 
which enables reasoning on incomplete data with the open-domain assumption.
Existential rules have the same logical form as database constraints known as tuple-generating dependencies, 
which have long been investigated \cite{AbiteboulHV}.  
Reborn under the names of existential rules, Datalog$^\exists$ or Datalog$^+$, they have raised significant interest in the last years as ontological languages, especially for the ontology-mediated query-answering and data-integration issues. 

\fussy

\medskip
A knowledge base (KB) is composed of a set of existential rules, which typically encodes ontological knowledge, 
and a factbase, which contains factual data. 
The forward chaining, also known as the \emph{chase} in databases, is a fundamental tool for reasoning on rule-based knowledge bases and a considerable literature has been devoted to its analysis. 
Its ubiquity in different domains comes from the fact it allows one to compute a \emph{universal model} of the  knowledge base, i.e., a model that maps by homomorphism to any other model of the knowledge base. 
This has a major implication in problems like answering queries with ontologies since
it follows that a (Boolean) conjunctive query is entailed by a KB if and only if it maps by homomorphism to a universal model.

Several variants of the chase have been defined: \emph{oblivious} or naive chase (e.g. \cite{cgk08}), \emph{skolem} chase  \cite{marnette09}, \emph{semi-oblivious} chase \cite{marnette09}, \emph{restricted} or standard chase \cite{fkmp05}, \emph{core} chase \cite{deutsch-nash-remmel08} (and its variant, the equivalent chase \cite{Rocher16}). 
All these chase variants compute logically equivalent results. \footnote{In addition, the \emph{parsimonious chase} was introduced in ~\cite{Leone:2012:ECD:3031843.3031846}. However, this chase variant, aimed towards responding at atomic queries, does not compute a universal model of the KB, hence it is outside the family of chase variants studied here.}
Nevertheless, they differ on their ability to detect the redundancies that are possibly caused by the introduction of unknown individuals (often called \emph{nulls}).
Note that, since redundancies can only be due to nulls, all chase variants output exactly the same results on rules without existential variables (i.e., Datalog rules, also called range-restricted rules \cite{AbiteboulHV}). 
Then, for rules with existential variables the chase produces iteratively new information until no new rule application is possible. The (re-)applicability of rules is depending on the ability of each chase variant to detect redundancies. Evidently this  has a direct impact on the termination. 
Of~course,~if~a~KB has no finite universal model then none of the chase variants will terminate. This is  illustrated by Example \ref{ex-init}. 
\begin{example}\label{ex-init}
Take the KB $\mathcal K = ( F, \mathcal R)$, where $\mathcal R$ contains the rule $R=\forall x\big( \textit{Human}(x) \rightarrow \exists y ~(\textit{parentOf}(y,x) \land \textit{Human}(y))\big)$  and $F = \{\textit{Human}(\textit{Alice})\}$. 
The application of the rule $R$ on the initial factbase $F$,
entails the existence of a new (unknown) individual $y_0$ 
(a \emph{null}) generated by the existential variable $y$ in the rule. 
This yields the factbase $\{\textit{Human}(\textit{Alice}), \textit{parentOf}(y_0,\textit{Alice}), \textit{Human}(y_0)\}$, which is logically translated into an existentially closed formula: $\exists y_0\big (\textit{Human}(\textit{Alice}) \land \textit{parentOf}(y_0,\textit{Alice}) \land \textit{Human}(y_0) \big)$. 
Then, $R$ can be applied again by mapping $x$ to $y_0$ thereby creating a new individual $y_1$. It is easy to see that in this case the forward chaining does not halt, as the generation of each new individual enables a novel rule application. 
This 
follows from the fact that the universal model of the knowledge base is infinite.
\hfill $\triangle$\vspace{0.2cm}\end{example}
However, for the case of KBs which have a \emph{finite} universal model, all chase variants  can be totally ordered with respect to the inclusion of the sets of factbases on which they halt: 
oblivious $<$ semi-oblivious = skolem $<$ restricted $<$ core. 
Here, $X_1 < X_2$ means that when $X_1$ halts on a KB, so does $X_2$, and there are KBs for which the reciprocal is false. 
The oblivious chase is the most redundant kind of the chase as it performs all possible rule applications, without checking for redundancies.
The core chase is the less redundant chase as it computes a minimal universal model by reducing every intermediate factbase to its core. 
In between, we find the semi-oblivious chase (equivalent to the skolem-chase) and the restricted chase. 
The first one does not consider isomorphic facts that would be generated by consecutive applications of a rule according to the same mapping of its frontier variables (i.e, variables shared by the rule body and head).  The second one discards all rule applications that produce ``locally redundant'' facts. The chase variants are illustrated by Example \ref{ex-chase-var}
 (for better presentation, universal quantifiers of rules will be omitted in the examples):

\sloppy
 \begin{example}\label{ex-chase-var}
Consider the knowledge bases $\mathcal K_1=(F,\{R_1\}), \mathcal
K_2=(F,\{R_2\})$, and $\mathcal K_3=(F',\{R_3\})$ built from the facts
  $F=\{\textit{p}(\textit{a},\textit{a})\}$ and
  $F'=\{\exists w ~\textit{p}(\textit{a},  w)\}$
    and the rules
  $R_1=\textit{p}(x,y){\rightarrow} \exists z  ~\textit{p}(x,z)$,
  $R_2=\textit{p}(x,y){\rightarrow} \exists z  ~\textit{p}(y,z)$ and
  $R_3=\textit{p}(x,y){\rightarrow} \exists z  ~(\textit{p}(x,x)\wedge\textit{p}(y,z))$.
Then, the oblivious chase does not halt~on~$\mathcal K_1$ while the
semi-oblivious chase does. Indeed, there are infinitely many different rule
applications on the atoms $\textit{p}(\textit{a},
z_0),$ $\textit{p}(\textit{a}, z_1),$ $\dots$ that can be generated with $
R_1$; yet, all rule applications map the frontier variable $x$ to the same
constant $a$, and are therefore filtered by the semi-oblivious chase.
In turn, the semi-oblivious chase does not halt on $\mathcal K_2$ while the
restricted chase does.
   Here again, there are infinitely many rule applications
on the atoms $\textit{p}(\textit{a}, z_0),\textit{p}( z_0, z_1),\dots$
that can be generated with $ R_2$; since each of them maps the frontier variables
    to new existentials, they are all performed by the semi-oblivious chase.
     However, all generated atoms are redundant with the initial atom
$\textit{p}(\textit{a},\textit{a})$   and the restricted chase deems the
first (and then all successive) rule applications as redundant.
  On the other hand, the restricted chase does not halt on $\mathcal K_3$ while the
core chase does. In this case, the first rule application yields
  $\exists  w \exists z_0 (\textit{p}(\textit{a},  w) \land
\textit{p}(\textit{a}, \textit{a}) \land \textit{p}( w, z_0))$. This
is logically
  equivalent~to~$\textit{p}(\textit{a},\textit{a})$
   i.e., its core, which leads to the core-chase termination at the next
step.
  However, the restricted chase checks only for redundancy of the newly added atoms with respect to the previous factbase, and does not take into account that the addition of new atoms can cause redundancies elsewhere in the factbase (in this example, the previous atom $p(a,w)$ together with the new atom $p(w,z_0)$ are redundant with respect to the new atom $p(a,a)$). So with the restricted chase, $R_3$ will be always applicable. 
Finally, note that $\textit{p}(\textit{a},\textit{a})$ is a (finite)
universal model for all knowledge bases $\mathcal K_1, \mathcal K_2,$ and $
\mathcal K_3$.
\hfill $\triangle$\vspace{0.2cm}\end{example}
\fussy


\noindent The termination problem, which asks whether for a given set of rules the chase will terminate on any factbase, is undecidable for all chase variants
 \cite{deutsch-nash-remmel08,blm10,GogaczM14}. 
Following previous work on Datalog, we study the related problem of \emph{boundedness} in a \emph{breadth-first} setting, i.e., the chase performs rule applications that correspond to a certain breadth-first level before any rule application that corresponds to a higher breadth-first level.
 Then, given a chase variant $\mathrm X$, we call a set of rules \emph{$\mathrm X$-bounded} if there is $k$ (called the bound) such that, for any factbase, the $\mathrm X$-chase stops after at most $k$ breadth-first steps. 
 Of course, since chase variants differ with respect to termination, they also differ with respect to boundedness.  


Boundedness
ensures several
 semantic properties. Indeed, if a set of rules is $\mathrm X$-bounded with $k$ the bound, then, for any factbase $F$, the saturation of $F$ at rank $k$ (i.e., the factbase obtained from $F$ after $k$ $\mathrm X$-chase breadth-first steps) is a universal model of the KB;  the reciprocal also holds true for the core chase.  Moreover, boundedness also ensures the UCQ-rewritability property (also called the finite unification set property \cite{blms11}): any (Boolean) conjunctive query $q$ can be rewritten using the set of rules  $\mathcal R$ into a (Boolean) union of conjunctive queries $Q$ such that for any factbase $F$, $q$ is entailed by $(F, \mathcal R)$ if and only if $Q$ is entailed by $F$. It follows that many interesting static analysis problems such as query containment under existential rules become decidable when a ruleset is bounded. Note that the conjunctive query rewriting procedure can be designed in a such a way that it terminates within $k$ breadth-first steps with $k$ the bound for the core chase \cite{LeclereMU16}. Finally, from a practical viewpoint, the degree of boundedness can be seen as a measure of the recursivity of a ruleset, and most likely, this is reflected in the actual number of breadth-first steps required by the chase for a given factbase or the query rewriting process for a given query, which is expected to be much smaller than the theoretical bound. 
As illustrated by Example \ref{ex-init}, the presence of existential  variables in the rules can make the universal model of a knowledge base infinite and so the ruleset unbounded, even for the core chase. However, the importance of the boundedness problem has been recognized already for rules without existential variables. Indeed, the problem has been first posed and studied  for Datalog, where it has been shown to be undecidable \cite{HillebrandKMV95,Marcin1999}. 
Example \ref{ex-datalog} illustrates some cases of bounded and unbounded rulesets in this setting.



\begin{example}\label{ex-datalog}
Consider the rulesets $\mathcal R_1=\{R\}$ and  $\mathcal R_2=\{R,R'\}$
where
$R= \textit{p}(x,y) \land \textit{p}(y,z) \rightarrow \textit{p}(x,z)$ and
$R'= \textit{p}(x,y) \land \textit{p}(u,z) \rightarrow \textit{p}(x,z)$.
The set $\mathcal R_1$  contains a single transitivity rule for the
predicate $p$.
This set is clearly unbounded as for any integer $k$ there exists a
factbase $F=\{p(a_i,a_{i+1})\ | \ 0\leq i<2^k  \}$ that requires  $k$ chase
steps.
On the other hand, $\mathcal R_2$ also contains a rule that joins
individuals on disconnected atoms. In this case, we have that $i)$ if $R$
generates some facts then $R'$ generates these same facts as well  and
$ii)$ $R'$ needs to be applied only at the first step, for any $F$, as it
does not produce any new atom at a later step. Therefore, $\mathcal R_2$ is
bounded with the bound $k=1$. Note that since these examples are in Datalog, the specificities
of the chase variants do not play any role.
\hfill $\triangle$\vspace{0.2cm}\end{example}

\noindent Finally, the next example illustrates boundedness for non-Datalog rules.

\sloppy
\begin{example}\label{ex-bounded-2}
Consider the ruleset $\mathcal R = \{p(x,y) \rightarrow \exists z (p(y,z)
\land p(z,y))\}$ and the fact $F = \{\textit{p(a,b)}\}$.
With all variants, the first chase step yields $F_1=\{p(a,b), p(b,z_0),
p(z_0,b)\}$. Then, two new rule applications are possible, which map
$p(x,y)$ to $p(b,z_0)$ and $p(z_0,b)$, respectively. The oblivious and
semi-oblivious chases will perform these rule applications and go on forever. Hence, the chase on $\mathcal R$ is not bounded for these two
variants. On the other hand, the restricted chase does terminate. It will not perform any of these rule
applications on $F_1$. Indeed, the first application would add the facts
$\{p(z_0,z_1), p(z_1,z_0)\}$, which can ``folded'' into $F_1$ by a homomorphism that maps 
$z_1$ to $b$ (while leaving $z_0$ fixed), and this is similar for the
second rule application.
We can check that actually the restricted chase will stop on any factbase,
and is bounded with $k = 1$. The same holds here for the core chase. 
\hfill $\triangle$\vspace{0.2cm}\end{example}
\fussy



%


\noindent Despite the relatively negative results on boundedness, knowing that a set of rules is bounded for some chase variant does not help much in practice anyway,  if the bound is unknown or even very large. 
Hence, the goal of this  paper is to investigate the $k$-boundedness problem, which asks, for a given chase variant, whether for any factbase, the chase stops after at most $k$ breadth-first steps. 

\medskip
Our main contribution is to show that $k$-boundedness is indeed decidable for the oblivious, semi-oblivious and restricted chases. 
Actually, we obtain a stronger result by exhibiting a property that a chase variant may fulfill, namely consistent heredity, and prove that $k$-boundedness is decidable as soon as this property is satisfied.  We show that it is the case for all the known chase variants except for the core chase. Hence, the decidability of   $k$-boundedness for the core chase remains an open question.  



\medskip


\section{Preliminaries } \label{sec:framework}

We consider a first-order setting with constants but no other function symbols. 
A term is either a constant or a variable. 
An atom is of the form $r(t_1,\dots, t_n)$ where $r$ is a predicate of
arity $n$ and the $t_i$ are terms. 
Given a set of atoms $A$, we denote by $\textit{vars}(A)  $ and $\textit{terms}(A)$ the set of its variables and terms.
A \emph{factbase} is a set of atoms, logically interpreted as the existentially closed conjunction of these atoms. 
A \emph{homomorphism} from a set of atoms $A$ to a set of atoms $B$ (notation: $\pi: A \rightarrow B$), is a substitution $\pi:\textit{vars}(A)\rightarrow \textit{terms}(B)$ such that $\pi(A)\subseteq B$. 
In this case, we also say that $A$ maps to $B$ (by $\pi$). \mlm{A homomorphism from $A$ to $B$ is an \emph{isomorphism} if its inverse is also a homomorphism. A set of atoms $A$ is a \emph{core} if there is no homomorphism from $A$ to one of its strict subsets. }
We denote by $\models$ the classical logical consequence and by $\equiv$ the logical equivalence. It is well-known that, given sets of atoms $A$ and $B$ seen as existentially closed conjunctions, there is a homomorphism from $A$ to $B$ if and only if $B \models A$.
\mlm{ When $A$ and $B$ are cores, $A \equiv B$ if and only if there is an isomorphism from $A$ to $B$. }

An \emph{existential rule} (or simply rule), denoted by $R$, is a formula $\forall \bar x \forall \bar y\big(B(\bar x,\bar y) \rightarrow \exists \bar z ~ H(\bar x, \bar z)\big)$
where $B$ and $H$, called the body and the head of the rule, 
are conjunctions of atoms, $\bar x$ and $\bar y$ are sets of universally quantified variables, and $\bar z$ is a set of existentially quantified variables. We call \emph{frontier} the variables  shared by the body and head of the rule, that is $\textit{frontier}(R)=\bar x$. 
In the following we will
refer to a rule as a pair of sets of atoms $(B, H )$ by interpreting their common variables as
the frontier. 
 A \emph{knowledge base} (KB) $\mathcal K = (F, \mathcal R)$ is a pair where $F$ is a factbase and $\mathcal R$ is a set of existential rules. We implicitly assume that all the rules as well as the factbase employ disjoint sets of variables, even if, for convenience, we reuse variable names in examples.

\medskip 

 Let $F$ be a factbase and $R=(B,H)$ be an existential rule. We say that $R$ is 
  \emph{applicable} on $F$ via $\pi$ 
  if there exists a homomorphism $\pi$ from its body $B$
  to $F$. We call the pair $(R,\pi)$ a \emph{trigger}.  
We denote by $\pi^s$ a \emph{safe} extension of $\pi$ which maps 
all existentially quantified variables in $H$ to fresh variables as follows : for each existential variable $z$  we have that $\pi^s(z)=z_{(R,\pi)}$\footnote{{This fixed way to choose a new fresh variable allows us to always produce the same atoms for a given trigger and that is without loss of generality since each trigger appears at most once on a derivation.}}.
The factbase $F\cup\pi^s(H)$ is called an \emph{immediate derivation} from $F$ through $(R,\pi)$.
Given a factbase $F$ and a ruleset  
$\mathcal{R}$ we define a \emph{derivation} from $F$ and $\mathcal{R}$, denoted by $\mathcal{D}$, as a (possibly infinite) sequence of triples 
$D_0=(\emptyset,\emptyset, F_0),D_1=(R_1,\pi_1, F_1),D_2=(R_2,\pi_2, F_2),$ $\dots$
 where $F_0=F$ and every $F_i$ $(i > 0)$ is an \emph{immediate derivation} from $F_{i-1}$ through a new trigger $(R_i,\pi_i)$, that is,
 $(R_i,\pi_i)\neq (R_j,\pi_j)$ for all $i\neq j$. 
   The \emph{sequence of rule applications 
   associated} with a derivation is simply the sequence of its triggers $(R_1,\pi_1),(R_2,\pi_2),\dots$  A \emph{subderivation} of a derivation $\mathcal{D}$ is any derivation $\mathcal{D}'$ whose sequence of rule applications is a subsequence\mlm{\footnote{A sequence $S$ is a subsequence of a sequence $S'$ if $S'$ can be obtained from $S$ by inserting some (or no) elements in $S$.}} of the sequence of rule applications associated with $\mathcal{D}$.


We will introduce four chase variants, namely oblivious (\OB), semi-oblivous (\SO), restricted (\RE), equivalent chase (\EQ). As explained later, some pairs of chase variants introduced in the literature have similar behavior, in which case we chose to focus on one of the two. 
All the chase variants are derivations that comply with some condition of applicability of the triggers. 

\begin{definition}\emph{Let $\mathcal{D}$ be a derivation of {length} $n$ from a factbase $F$ and a ruleset $\mathcal{R}$, and $F_n$ the factbase obtained after the $n$  rule applications of $\mathcal{D}$. A trigger $(R,\pi)$ is called: } \begin{enumerate}
\item \emph{\OB-\emph{applicable} on $\mathcal{D}$ if $R$ is applicable on $F_n$ via $\pi$.}
\item \emph{\SO-\emph{applicable} on $\mathcal{D}$ if $R$ is applicable on $F_n$ via $\pi$ and for every trigger $(R,\pi')$ in the sequence of triggers associated with $\mathcal{D}$, the restrictions of $\pi$ and $\pi'$ to the frontier of $R$ are not equal.}
\item \emph{\RE-\emph{applicable} on $\mathcal{D}$ if $R=(B,H)$ is applicable on $F_n$ via $\pi$ and $\pi$ cannot be extended to a homomorphism $\pi' : B\cup H\rightarrow F_n$.}
\item \emph{\EQ-\emph{applicable} on $\mathcal{D}$ if $R=(B,H)$ is applicable on $F_n$ via $\pi$ and it does not hold that $F_n\equiv F_n\cup \pi^s(H)$.}\eod\end{enumerate} \end{definition}

\noindent Note that for $ \mathrm{X} \in \{ \mathbf{O}, \mathbf{R}, \mathbf{E} \}$, the applicability of the trigger only depends on $F_n$ (hence we can also say the trigger is $\mathrm{X}$-\emph{applicable} on $F_n$), while for the $\mathbf{SO}$-chase we have to take into account the previous triggers. 
Note also that the definitions of \OB- and \SO- trigger applicability allow one  to extend a derivation with a rule application that does not add any atom, i.e., $F_{n+1}=F_{n}$; however, this is not troublesome since no derivation can contain twice the same triggers.


\medskip

Given a derivation $\mathcal{D}$,
we define the \emph{rank} of an atom as follows: 
$\textit{rank}(A)=0$ if $A\in F_0$, otherwise 
let $R=(B,H)$ and $(R,\pi)$  be the first trigger in the sequence $\mathcal D$ 
such that $A\in \pi^s(H)$, then
$\textit{rank}(A)=1+\max_{A'\in \pi(B)}\{\textit{rank}(A')\}$.
\mlm{When we consider a breadth-first chase, the rank of an atom intuitively corresponds  to the chase step at which it has been generated. }
This notion is naturally extended to triggers: $\textit{rank}((R,\pi))=1+ \max_{A'\in\pi(B)}\{\textit{rank}(A')\}$.
%
%
%

\medskip
The \emph{depth} of a finite derivation is the maximal rank of one of its atoms. 
Finally, a derivation $\mathcal D$ is $\mathrm{X}$-\emph{breadth-first} (where $\mathrm{X}\in\{\OB,\SO,\RE,\EQ\}$) if it satisfies the following two properties:\begin{itemize} \item\emph{(1) rank compatibility:} for all elements $D_i$ and $D_j$ in $\mathcal D$ with $i <j$, 
the rank of the trigger of $D_i$ is smaller or equal to the rank of the trigger of $D_j$, and \item\emph{(2) rank exhaustiveness:} for every rank $k$ of a trigger in $\mathcal D$, let $D_i = (R_i, \pi_i, F_i)$ be the last element in $\mathcal D$ such that $rank ((R_i, \pi_i)) = k$. Then, every trigger which is $\mathrm{X}$-applicable on the subderivation $D_1,...,D_i$ is of rank $k+1$. 
\end{itemize} 



\begin{definition}[Chase variants]
\emph{Let $F$ be a factbase and $\mathcal{R}$ be a ruleset. We define four \emph{variants} of the chase:}
\begin{enumerate}
\item \emph{An \textbf{oblivious chase} is any derivation $\mathcal D$ from $F$ and $\mathcal{R}$.}
\vspace{2mm}
\item \emph{A \textbf{semi-oblivious chase} 
is any derivation $\mathcal D$ from $F$ and $\mathcal R$ such that for every element $D_i=(R_i,\pi_i,F_i)$ of $\mathcal{D}$, the trigger $(R_i,\pi_i)$ is \SO-applicable on
the subderivation $D_0,D_1,...,D_{i-1}$ of $\mathcal{D}$.}

\vspace{2mm}
\item \emph{A \textbf{restricted chase} 
is any derivation $\mathcal D$ from $F$ and $\mathcal R$ such that for every element $D_i=(R_i,\pi_i,F_i)$ of $\mathcal{D}$, the trigger $(R_i,\pi_i)$ is \RE-applicable to on
the subderivation $D_0,D_1,...,D_{i-1}$ of $\mathcal{D}$.}

\vspace{2mm}
\item \emph{An \textbf{equivalent chase} is any \EQ-breadth-first  derivation $\mathcal D$ from $F$ and $\mathcal R$ such that for every element $D_i=(R_i,\pi_i,F_i)$ of $\mathcal{D}$, the trigger $(R_i,\pi_i)$ is \EQ-applicable on
the subderivation $D_0,D_1,...,D_{i-1}$ of $\mathcal{D}$.}\eod
\end{enumerate}
\end{definition} 

\noindent
We will abbreviate the above chase variants with 
$\mathbf{O}$-chase, 
$\mathbf{SO}$-chase, 
$\mathbf{R}$-chase, and 
$\mathbf{E}$-chase, respectively. Unless otherwise specified, when we use the term $\mathrm{X}$-chase derivation, we will be referring to any of the four chase variants.
Furthermore, with \emph{breadth-first} X-chase derivation we will always imply X-breadth-first X-chase derivation.

\medskip
An X-chase derivation $\mathcal{D}$ from $F$ and $\mathcal{R}$ is \emph{exhaustive} if for all $i\geq 0$, if a trigger $(R,\pi)$ is X-applicable on the subderivation $D_1,...,D_i$, then there is a $k\geq i$ such that one of the two following holds:
\begin{enumerate}
\item ${D}_k=(R,\pi,F_k)$ or
\item $(R,\pi)$ is not X-applicable on $D_1,...,D_k$.
\end{enumerate}

\noindent\mlm{Exhaustivity is also known as \emph{fairness}}. An X-chase derivation is  \emph{terminating} if it is both exhaustive and finite. 


It is well-known that for $ X \in \{ \mathbf{O}, \mathbf{SO}, \mathbf{E} \}$, if there exists a terminating derivation for a given KB, then all exhaustive derivations on this KB are terminating. This does not hold for the restricted chase, because the order in which rules are applied matters, as illustrated by the next example:

\begin{example}\label{restterm} We assume two rules $R_1= p(x,y) \rightarrow \exists z ~p(y,z)$ and $R_2= p(x,y) \rightarrow p(y,y)$ and $F = \{p(a,b)\}$. Let $\pi=\{x\mapsto a, y\mapsto b\}$. Then $(R_1,\pi)$ and $(R_2,\pi)$ are both \RE-applicable. If $(R_2,\pi)$ is applied first, then the derivation is terminating. However if we apply $(R_1,\pi)$ first, and $(R_2,\pi)$ second we produce the factbase $F_2=\{p(a,b),p(b,z_{(R_1,\pi)}),p(b,b)\}$ and with $\pi'=\{x\mapsto b, y\mapsto z_{(R_1,\pi)}\}$ we have that $(R_1,\pi')$ as well as $(R_2,\pi')$ are again both \RE-applicable. Consequently, if we always choose to apply $R_1$ before $R_2$ then the corresponding derivation will be infinite.
\hfill $\triangle$\vspace{0.2cm}\end{example}

\medskip \mlm{Let us now link the four previous chase variants to some other known chase variants.} The semi-oblivious and skolem chases, both defined in \cite{marnette09}, lead to similar derivations. Briefly, the skolem chase consists of first skolemizing the rules (by replacing existentially quantified variables with skolem functions whose arguments are the frontier variables) then running the oblivious chase. Both chase variants yield isomorphic results, in the sense that they generate exactly the same sets of atoms, up to a bijective renaming of nulls by skolem terms. Therefore, we chose to focus on one of the two, namely the semi-oblivious chase. The core chase \cite{deutsch-nash-remmel08} and the equivalent chase \cite{Rocher16} have similar behaviors as well. We remind that a core of a set of atoms is one of its minimal equivalent subsets, \mlm{and that two equivalent sets of atoms have isomorphic cores}. The core chase proceeds in a breadth-first manner and, at each step, performs in parallel all rule applications according to the restricted chase criterion, then computes a core of the resulting factbase.  Hence, the core chase may remove at some step atoms that were introduced at a former step. 
\mlm{After $i$ breadth-first steps, the equivalent chase and the core chase yield logically equivalent factbases, and they terminate on the same inputs.  
This follows from the facts that computing the core after each rule application or after a sequence of rule applications gives isomomorphic results, and that $F_i\equiv F_{i+1}$ if and only if $\textit{core}(F_i)$ is isomorphic to $\textit{core}(F_{i+1})$.} However, it is sometimes more convenient to handle the equivalent chase from a formal point of view because of its monotonicity (in the sense that within a derivation $F_i\subseteq F_{i+1}$).

\noindent We now introduce some notions that will be central for establishing our results on $k$-boundedness for the different chase variants.

\begin{definition}[Restriction of a derivation]\emph{
Let $\mathcal{D}$ be a derivation from $F$ and $\mathcal R$.
For any $G \subseteq F$,  the \emph{restriction of $\mathcal{D}$ induced by} $G$
denoted by $\mathcal{D}_{|G}$, 
is the maximal derivation from $G$ and $\mathcal{R}$ obtained by a subsequence of the trigger sequence of $\mathcal{D}$. 
}\eod
\end{definition}


\noindent The following example serves to demonstrate how a subset of the initial factbase induces the restriction of a derivation:
\begin{example}
Take $F=\{\textit{p}(\textit{a},\textit{a}),\textit{p}(\textit{b},\textit{b})\}$, 
$R=p(x,y)\rightarrow \exists z~ p(y,z)$ and 
\[\mathcal D=(\emptyset,\emptyset,F),(R,\pi_1,F_1),(R,\pi_2,F_2),(R,\pi_3,F_3),(R,\pi_4,F_4)\] with 
$\pi_1=\{x/y\mapsto a\}$,
$\pi_2=\{x/y\mapsto b\}$,
$\pi_3=\{x\mapsto a,y\mapsto z_{(R,\pi_1)}\}$, and
${\pi_4=\{x\mapsto z_{(R,\pi_1)},y\mapsto z_{(R,\pi_3)}\}}$. 

\medskip
\noindent The derivation $\mathcal D$ produces the factbase 
\[F_4~{=}~F\cup\{
\textit{p}(\textit{a},z_{(R,\pi_1)}),
\textit{p}(\textit{b},z_{(R,\pi_2)}),
\textit{p}(z_{(R,\pi_1)},z_{(R,\pi_3)}),
\textit{p}(z_{(R,\pi_3)},z_{(R,\pi_4)})
\}\] 
Then, if $G=\{p(a,a)\}$, we have
$\mathcal D_{|G}=(\emptyset,\emptyset,G),(R,\pi_1,G_1),(R,\pi_3,G_2),(R,\pi_4,G_3)$
is the restriction of $\mathcal D$ induced by $G$ where \[G_3=G\cup\{\textit{p}(\textit{a},z_{(R,\pi_1)}),\textit{p}(z_{(R,\pi_1)},z_{(R,\pi_3)}),\textit{p}(z_{(R,\pi_3)},z_{(R,\pi_4)})\}\]
\hfill $\triangle$\vspace{0.2cm}\end{example}

\begin{definition}[Ancestors]\emph{
\noindent Let $D_i=(R_i,\pi_i,F_i)$ be an element of a derivation $\mathcal{D}$. Then every atom in $\pi_i(B_i)$ is called a \emph{direct ancestor} of every atom in $(F_i \setminus F_{i-1})$. The (indirect) \emph{ancestor} relation between atoms is defined as the transitive closure of the direct ancestor relation.
The direct and indirect ancestor relations between atoms are extended to triggers: let $D_j=(R_j,\pi_j,F_j)$ where $j<i$. Then $(R_j,\pi_j)$ is a \emph{direct ancestor} of $(R_i,\pi_i)$ if there is an atom in $(F_j\setminus F_{j-1})$ which is a direct ancestor of the atoms in $(F_i\setminus F_{i-1})$.
We will denote the ancestors of sets of atoms and triggers as $\textit{Anc}(F,\mathcal{D})$ and $\textit{Anc}((R,\pi),\mathcal{D})$, respectively. 
The inverse of the ancestor relation is called the \emph{descendant} relation.} \eod\end{definition}

\noindent There is an evident correspondence between the notion of ancestors and the notions of rank and depth. Suppose a ruleset with at most $b$ atoms in the rules' bodies. The following lemma results from the fact that each atom has at most $b$ direct ancestors and \mlm{the length of} a chain of ancestors cannot exceed the depth of a derivation.

\begin{lemma}[The ancestor clue]\label{anclue}\emph{Let $\mathcal{D}$ be an $X$-chase derivation from $F$ and $\mathcal R$. Then for any atom $A$ of rank $k$ in $\mathcal{D}$, $|F\cap Anc\big(A,{\mathcal{D}}\big) |\leq b^k$; also for any trigger $(R,\pi)$ of rank $k$ in $\mathcal{D}$, $|F\cap Anc\big((R,\pi),{\mathcal{D}}\big) |\leq b^k$.}
\end{lemma}

\noindent This lemma will be instrumental for proving our results on $k$-boundedness as it allows one to characterize the \mlm{maximal} number of atoms that are needed to produce a new atom at a given chase step.

In the next section, we turn our attention to the properties of the derivations that are key to study $k$-boundedness.

\section{Breadth-first Boundedness}

As already mentioned, the concept of boundedness was first introduced for Datalog programs. A Datalog program is said to be bounded if the number of breadth-first steps of a bottom-up evaluation of the program is bounded independently from any database (this notion being more precisely called \emph{uniform boundedness} to distinguish it from the notion of \emph{program boundedness} that restricts the set of predicates that may occur in the database)~\cite{GaifmanMSV93,Abiteboul89,GuessarianV94}. 
Applying this concept to the more general language of existential rules, and parametrizing it by the considered chase variant, $\mathrm X$-boundedness can be specified as follows: 

\begin{definition}
\emph{Let  $\mathrm{X}\in\{\mathbf{O},\mathbf{SO},\mathbf{R},\mathbf{E}\}$. A ruleset $\mathcal R$ is $\mathrm{X}$-bounded if there is $k\in\mathbb N$ such that for every factbase $F$, every breadth-first $\mathrm{X}$-chase derivation is of depth at most $k$. \eod}
\end{definition}

\noindent This definition may seem natural, however it deserves some comments. First note that in Datalog all exhaustive derivations have the same length but not necessarily the same depth, as illustrated by the following example.


\begin{example}
Let $F=\{p(a)\}$ and $\mathcal R=\{R_1,R_2,R_3\}$ where $R_1=p(x)\rightarrow q(x)$, $R_2=q(x)\rightarrow r(x)$, $R_3=p(x)\rightarrow r(x)$. Here are two exhaustive derivations:
\[ \mathcal D_1 = (\emptyset,\emptyset,F), (R_1,\pi,F_1), (R_2,\pi,F_2), (R_3,\pi,F_2)\]
\[ \mathcal D_2 = (\emptyset,\emptyset,F), (R_1,\pi,F_1), (R_3,\pi,F_2), (R_2,\pi,F_2)\]
where $\pi=\{x\mapsto a\}$. We can see that both derivations are exhaustive, however the depth of $\mathcal D_1$ is 2 whereas the depth of $\mathcal D_2$ is 1.
\hfill $\triangle$
\end{example}

\noindent However, among all exhaustive derivations with Datalog rules, the class of breadth-first derivations are of minimal depth. This remains true for the oblivious and semi-oblivious chase derivations with existential rules:

\begin{proposition}\label{res}\emph{
For each terminating $\mathbf{O}$-chase derivation  (resp. $\mathbf{SO}$-chase derivation)  from $F$ and $\mathcal R$
there exists a breadth-first terminating $\mathbf{O}$-chase derivation  (resp. $\mathbf{SO}$-chase derivation) from  $F$ and $\mathcal R$ 
of smaller or equal depth.} 
\end{proposition}  

\begin{proofi}
If $\mathcal D$ is a terminating \OB-chase derivation, we can \mlm{reorder} the sequence of triggers associated with $\mathcal D$ in such a way as to create a rank compatible \OB-chase derivation $\mathcal D'$ (we know that the applicability condition is not affected if we perform some rule applications earlier). Then $\mathcal D'$ is also exhaustive since the resulting factbase is the same. Moreover $\mathcal D'$ has to be rank exhaustive, since if a trigger is \OB-applicable on a factbase at some step of the derivation, it is always \OB-applicable (unless it has already been applied). So $\mathcal D'$ is breadth-first. 

Let us now consider \SO-chase derivations. For convenience in the following proof, given a trigger $(R,\pi)$, we slightly modify the definition of the safe extension $\pi^s$: for each existential variable $z$ in $H$ (the head of $R$), we define $\pi^s(z) = z_{f_R(\pi(x_1), ...\pi(x_n)})$ where
$f_R$ is a fresh symbol assigned to $R$, and $(x_1,...,x_n)$ is a fixed ordering of the frontier variables in $R$. For brevity, we say that two triggers $(R,\pi)$ and $(R,\pi')$ such that $\pi$ and $\pi'$ have the same restriction to the frontier of $R$ are ``frontier-equal''. With the new definition, two frontier-equal triggers produce exactly the same set of atoms, i.e., $\pi^s(H) = \pi'^s(H)$. Since a \SO-chase derivation does not have frontier-equal triggers, this modification of the names of fresh variables can be done without loss of generality. 

Let $\mathcal D$ be a terminating \SO-chase derivation from a factbase $F$. 
We build a derivation $\mathcal D_{bf}$ from $\mathcal D$ by increasing rank as follows. Let 
$\mathcal D_0 = \mathcal D \setminus (\emptyset, \emptyset, F)$, $\mathcal D_{bf}^0 = (\emptyset, \emptyset, F)$. 
Starting from $i=1$, we iteratively perform the following steps:\\
1) Let $T$ be the set of all triggers $(R,\pi)$ from $\mathcal D_{i-1}$ such that there is a frontier-equal trigger $(R,\pi')$ applicable on $\mathcal D_{bf}^{i-1}$, and let $T'$ be the set composed of one trigger $(R,\pi')$ for each $(R,\pi)$ in $T$.\\
2) If $T = \emptyset$, $\mathcal D_{bf} = \mathcal D_{bf}^{i-1}$. \\
3) Otherwise, $\mathcal D_{bf}^i$ is obtained by extending $\mathcal D_{bf}^{i-1}$ with the triples corresponding to the triggers in $T'$ (in any order), and $\mathcal D_{i}$ is obtained from $\mathcal D_{i-1}$ by removing the triples corresponding to the triggers in $T$.

We can easily check that the following conditions are fulfilled at each step of the algorithm: (a)  $\mathcal D_{bf}^i . \mathcal D_i$ is a well-formed derivation (b) there is a bijection between the triggers in $\mathcal D$ and those in $\mathcal D_{bf}^i . \mathcal D_i$, such that corresponding triggers are frontier-equal; (c) the depth of $\mathcal D_{bf}^i$ is less or equal to the depth of  $\mathcal D$; (d) $\mathcal D_{bf}^i$ is a breadth-first derivation. For Point (a), note that replacing  $(R,\pi)$ by $(R,\pi')$ has no impact on the name of the obtained fresh variables, hence no impact on triggers that use atoms produced  by $(R,\pi)$. For Point (d), note that $\mathcal D_{bf}^i$ is rank-compatible by construction, and that it is rank-exhaustive: otherwise, there would be a trigger $(R,\pi)$ still \SO-applicable on $\mathcal D$, which is not possible since $\mathcal D$ is terminating. 

The algorithm terminates since the number of steps is upper bounded by the depth of $\mathcal D$. Let $i=d$ be the last step. Then, $\mathcal D_{d-1} = \emptyset$, hence, from (b),  there is a bijection between the triggers in $\mathcal D$ and those in $\mathcal D_{bf} = \mathcal D_{bf}^{d-1}$, such that corresponding triggers are frontier-equal. It follows that $\mathcal D_{bf}$ is terminating. 
\end{proofi}


\noindent The equivalent chase, which is inspired from the core chase, is breadth-first by definition. The case of the restricted chase is more complex, since, \mlm{for a given factbase}, some exhaustive derivations may terminate, while others may not. It may happen that all breadth-first derivations terminate (with depth less than a predefined number $k$), but there is an exhaustive non-breadth-first derivation that does not terminate. It may also be the case that no breadth-first derivation terminates, but there is a non-breadth-first derivation that terminates (with predefined depth less than $k$), as illustrated by the next example.

\begin{example} Let $F=\{p(a,b)\}$ and $\mathcal R=\{R_1,R_2,R_3\}$ with $R_1=p(x,y)\rightarrow \exists z \ p(y,z)$, $R_2=p(x,y)\rightarrow\exists z\  q(y,z)$ and $R_3=q(y,z)\rightarrow p(y,y)$. It is easy to see that a breadth-first \RE-chase derivation in this knowledge base cannot be terminating. However by applying only $R_2$ on $F$ and then $R_3$ on the new atom, we obtain a terminating \RE-chase derivation. \mlm{Note also that, for any factbase, there is a terminating \RE-chase derivation of depth at most 2.}
\hfill $\triangle$\end{example}

\noindent Hence, in the case of the restricted chase, breadth-first derivations may not be derivations of minimal depth. More generally, one cannot exclude that other classes of derivations behave better with respect to depth. Moreover, it would be interesting to parametrize boundedness with respect to a specific kind of derivation that would be computed by some restricted chase algorithm.  Therefore, a more general definition of boundedness could be based on the maximal depth of a class of derivations of interest. Then, boundedness based on breadth-first settings, as studied in this paper, could be seen as depth-based boundedness applied to breadth-first X-chase variants.  

\medskip
Finally, the following property gives more insight on the relationships between $\RE$-chase derivations and rank-compatible $\RE$-chase derivations
(we recall that breadth-first derivations are rank-compatible derivations that are moreover rank-exhaustive).

\begin{proposition}\label{res}\emph{
For each terminating $\mathbf{R}$-chase derivation    from $F$ and $\mathcal R$
there exists a  terminating rank-compatible $\mathbf{R}$-chase derivation   from  $F$ and $\mathcal R$ 
of smaller or equal depth.} 
\end{proposition}

\begin{proofi}  \sloppy
Let $\mathcal{D}$ be a terminating $\mathbf{R}$-chase derivation from $F$ and $\mathcal{R}$. 
Let $\mathcal{T}_{\mathcal{D}}$ be its sequence of associated triggers and let $\mathcal{T}$ be a sorting of $\mathcal{T}_{\mathcal{D}}$ such that the rank of each element is greater or equal to the rank of its predecessors. Note that $\mathcal T$ contains exactly the same triggers as $\mathcal T_{\mathcal D}$, only the order has changed.
Let $\mathcal{D}'$ be the derivation defined by applying, when $\mathbf{R}$-applicable, the triggers using the order of $\mathcal{T}$. 
Because of the reordering, some of the triggers in $\mathcal{T}$ may no longer be $\mathbf{R}$-applicable in $\mathcal{D}'$. However, $\mathcal{D}'$ respects the rank compatibility property. We will show that it is a terminating $\RE$-chase derivation. 
Suppose that there is a new trigger $(R,\pi)$ (not present in $\mathcal{T}$) which is $\mathbf{R}$-applicable on $\mathcal{D}'$ (with $R=(B,H)$). Let $\hat F$ be the resulting factbase from $\mathcal D'$. So we can say that $(R,\pi)$ is \RE-applicable on $\hat F$. Let $\tilde F$ be the resulting factbase from $\mathcal D$.
Then, since $\hat F\subseteq \tilde F$, we have that $(R,\pi)$ is $\mathbf{O}$-applicable on $\tilde F$. But because $\mathcal{D}$ is a terminating \RE-chase derivation, we know that $(R,\pi)$ in not \RE-applicable on $\tilde F$. Let $(R_1,\pi_1),...,(R_m,\pi_m)$ be
the triggers of $\mathcal{T}_{\mathcal{D}}$ that do not appear in  $\mathcal{D}'$ (i.e., were not \RE-applicable when constructing $\mathcal{D}'$). So \begin{equation}\label{thisequ}\tilde F=\hat F\cup \pi_1^s(H_1)\cup\cdots\cup\pi_m^s(H_m)\end{equation} where $H_1,...,H_m$ are the heads of the rules $R_1,...,R_m$ respectively.
Since $(R,\pi)$ is not \RE-applicable on $\tilde F$ we conclude that there is a homomorphism from $\pi^s(H)$ to $\tilde F$, i.e., a substitution $\sigma:\vars{\pi^s(H)}\rightarrow\terms{\tilde F}$ such that $\sigma(\pi^s(H))\subseteq \tilde F$,   
while $\sigma$ is the identity on $\pi(B)$. Since $(R_1,\pi_1),...,(R_m,\pi_m)$ are not \RE-applicable in $\mathcal D'$ we know that there are substitutions $\sigma_1,...,\sigma_m$ such that for every $i\in\{1,...,m\}$ we have $\sigma_i:\vars{\pi_i^s(H_i)}\rightarrow\terms{\hat F}$ and $\sigma_i(\pi_i^s(H_i))\subseteq \hat F$ (i.e., homomorphisms from $\pi_i^s(H_i)$ to $F$),  where $\sigma_i$ is the identity on $\pi_i(B_i)$.
Since with $\sigma_1,...,\sigma_m$, only new variables are mapped to different terms (and all other variables are mapped to themselves), we can define the substitution $\displaystyle\dot\sigma=\bigcup_{i=1}^m \sigma_i$ which has the property that \begin{equation}\label{dyo}\dot\sigma\big(\hat F\cup \pi_1^s(H_1)\cup\cdots\cup\pi_m^s(H_m)\big)=\hat F\end{equation} Moreover, the set of variables that are not identically mapped from $\dot\sigma$ is disjoint with the variable set $\vars{\hat F}$, because the new variables created from $(R_1,\pi_1),...,(R_m,\pi_m)$ are not present in $\hat F$. Therefore the composition $\dot\sigma\circ\sigma$ retains the set of new variables in $\pi^s(H)$ as its set of variables mapped to different terms. So by~\ref{thisequ} and $\sigma(\pi^s(H))\subseteq \tilde F$ we can write \[\dot\sigma\circ\sigma\big(\pi^s(H)\big)\subseteq\dot\sigma\big(\hat F\cup \pi_1^s(H_1)\cup\cdots\cup\pi_m^s(H_m)\big)\] which with~\ref{dyo} becomes
\[\dot\sigma\circ\sigma\big(\pi^s(H)\big)\subseteq \tilde F\] which implies that $(R,\pi)$ is not \RE-applicable on $\mathcal D'$. That is a contradiction, which leads us to conclude that no such $(R,\pi)$ exists, therefore $\mathcal D'$ is a terminating \RE-chase derivation.
\end{proofi}

\noindent As already mentioned, boundedness is shown to be undecidable for classes of existential rules like Datalog. However, the practical interest of this notion lies more on whether we can find the particular bound $k$, rather than knowing that there exists one and thus the ruleset is bounded. 
Because even if we cannot know whether a ruleset is bounded or not, it can be useful to be able to check a particular bound $k$.  
To this aim, we define the notion of $k$-boundedness where the bound is known, and we prove its decidability for three of the four chase variants.

\section{Decidability of $k$-boundedness for some chase variants}


\begin{definition}[$k$-boundedness]\emph{Given a chase variant $\mathrm{X}$, a ruleset $\mathcal R$ is $\mathrm{X}$-\emph{$k$-bounded} if for every factbase $F$, every breadth-first $\mathrm{X}$-chase derivation is terminating with depth at most $k$.}\eod
\end{definition}

\noindent Note that  a ruleset which is $k$-bounded is also bounded, but the converse is not true. 
Our approach for testing $k$-boundedness is to construct a finite set of factbases whose size depends solely on $k$ and $\mathcal R$, that acts as representative of \emph{all} factbases for the boundedness problem.
From this one could obtain the decidability of $k$-boundedness.
Indeed, for each representative factbase one can compute all breadth-first derivations of depth $k$ and check if they are terminating. 
 
For analogy, it is well-known that the oblivious chase terminates on all factbases if and only if it terminates on the so-called critical instance
(i.e., the instance that contains all possible atoms on the constants occurring in rule bodies, with a special constant being  chosen if the rule bodies have only variables)~\cite{marnette09}.  However, it can be easily checked that the critical instance does not provide oblivious chase derivations of maximal depth, hence is not suitable for our purpose of testing $k$-boundedness.
Also, to the best of our knowledge, no representative sets of all factbases are known for the termination of the other chase variants. 

In this section, we prove that $k$-boundedness is decidable for the oblivious, semi-oblivious (skolem) and restricted chase variants by exhibiting such representative factbases. 
A common property of these three chase variants is that redundancies can be checked ``locally'' within the scope of a rank, while in the equivalent chase, redundancies may be ``global'', in the sense by adding an atom we can suddenly make redundant atoms added by previous ranks. 

Following this intuition, we define the notion of hereditary chase.

\begin{definition}\emph{The $\mathrm{X}$-chase is said to be \emph{hereditary} if, for any   $\mathrm{X}$-chase derivation $\mathcal{D}$ from $F$ and $\mathcal R$, the restriction of $\mathcal{D}$ induced by $F' \subseteq F$ is an $\mathrm{X}$-chase derivation. }\eod
\end{definition}
A chase is hereditary if by restricting a derivation on a subset of a factbase we still get a derivation with no redundancies. This captures the fact that redundancies can be tested ``locally''.
 This property is fulfilled by the oblivious, semi-oblivious and restricted chase variants; a counter-example for the equivalent chase is given as the end of this section.

\begin{proposition}\emph{The $\mathrm{X}$-chase is hereditary for $\mathrm{X}\in\{\mathbf{O},\mathbf{SO},\mathbf{R} \}$.}  
\end{proposition}

\begin{proofi} 
We assume that $\mathcal{D}$ is an $X$-chase derivation from $F$ and $\mathcal{R}$, and $\mathcal{D}_{|F'}$ is the restriction of $\mathcal{D}$ induced by $F' \subseteq F$.

\noindent \textbf{Case O}  By definition, an $\mathbf{O}$-chase derivation is any sequence of immediate derivations with distinct triggers, so the restriction of a derivation from a subfact of $F$ is  an $\mathbf{O}$-chase  derivation.

\noindent \textbf{Case SO} The condition for $\mathbf{SO}$-applicability is that we do not have two triggers which map frontier variables in the same way.
 As  $\mathcal{D}$ fulfills this condition its subsequence $\mathcal{D}_{|F'}$ also fulfills it.

\noindent \textbf{Case R} The condition for $\mathbf{R}$-applicability imposes that for a trigger $(R,\pi)$ there is no extension of $\pi$ that maps the head of $R$ to $F$. Since $\mathcal{D}_{|F'}$ generates a factbase included in the factbase generated by $\mathcal{D}$ we conclude that $\mathbf{R}$-applicability is preserved.
\end{proofi}

\noindent Note however that when  $\mathcal{D}$ is breadth-first, it not ensured that its restriction induced by $F'$ is still breadth-first (because the rank exhaustivness might not be satisfied). It is actually the case for the oblivious chase (since all triggers are always applied), but not for the other variants since some rule applications that would be possible from $F'$ have not been performed in $\mathcal{D}$ because they were redundant in $\mathcal{D}$ given the whole $F$. The next examples illustrate these cases. 

\medskip

\begin{example} [Semi-oblivious chase] 

\noindent Let $F = \{p(a,b), r(a,c)\}$ and $\mathcal R = \{R_1= p(x,y) \rightarrow r(x,y); R_2= r(x,y) \rightarrow \exists z~q(x,z); R_3=r(x,y) \rightarrow t(y)\}$. Let $\mathcal{D}$ be the (non terminating) breadth-first derivation of depth $2$ from $F$ whose sequence of associated triggers is $(R_1, \pi_1), $ $(R_3,\pi_2), $ $(R_2,\pi_2),(R_3,\pi_1)$ with $\pi_1=\{x\mapsto a, y\mapsto b\}$ and $\pi_2=\{x\mapsto a, y\mapsto c\}$ which produces $r(a,b),  t(c), q(a,z_{(R_2,\pi_2)}), t(b)$; the trigger $(R_2,\pi_1)$ is then O-applicable but not SO-applicable, as it maps equally the frontier variables as $(R_2, \pi_2)$. Let  $F' =\{ p(a,b)\}$. The restriction of $\mathcal{D}$ induced by $F'$ includes only $(R_1,\pi_1), (R_3,\pi_1)$ and is a SO-chase derivation of depth $2$, however it is not breadth-first since now $(R_2,\pi_1)$ is SO-applicable at rank $2$ (thus the rank exhaustiveness is not satisfied). 
\hfill $\triangle$\vspace{0.2cm}
\end{example}

\begin{example} [Restricted chase] 

\noindent
Let $F = \{p(a,b), q(a,c)\}$ and $\mathcal R = \{R_1=p(x,y) \rightarrow r(x,y); R_2=r(x,y) \rightarrow \exists z~q(x,z); R_3=r(x,y) \rightarrow t(x)\}$. Let $\mathcal{D}$ be the (terminating) breadth-first derivation of depth $2$ from $F$ whose sequence of associated triggers is $(R_1,\pi), (R_3,\pi)$ with $\pi=\{x\mapsto a, y\mapsto b\}$ which produces $\{p(a,b), q(a,c), r(a,b), t(a)\}$; note that the trigger $(R_2,\pi)$ is SO-applicable but not R-applicable because of the presence of $q(a,c)$ in $F$. Let  $F' = \{p(a,b)\}$. The restriction of $\mathcal{D}$ induced by $F'$ is a restricted chase derivation of depth $2$, however it is not breadth-first since now $(R_2,\pi)$ is R-applicable at rank $2$ and thus has to be applied (to ensure the rank exhaustiveness of a breadth-first derivation). 
\hfill $\triangle$\vspace{0.2cm}
\end{example}


\noindent \mlm{Previous examples illustrate the need for a more appropriate property focusing on breadth-first derivations. Hence, we define another property, namely \emph{consistent heredity}, which ensures that the restriction of a breadth-first derivation $\mathcal{D}$ induced by $F'$ \emph{can be extended}  to a breadth-first derivation (still from $F'$). 
When we consider breadth-first X-chases, heredity implies consistent heredity. }


\begin{definition} \emph{
The $\mathrm{X}$-chase is said to be \emph{consistently hereditary} if for any factbase $F$ and any breadth-first $\mathrm{X}$-chase derivation $\mathcal{D}$ from $F$ and $\mathcal{R}$, the restriction of $\mathcal{D}$ induced by $F' \subseteq F$ is a subderivation
 of a breadth-first $\mathrm{X}$-chase derivation $\mathcal{D}'$ from $F'$ and $\mathcal{R}$.}\eod \end{definition}

\begin{proposition}\label{conher}\emph{The $\mathrm{X}$-chase is consistently hereditary for $\mathrm{X}\in\{\mathbf{O},\mathbf{SO},\mathbf{R}\}$.}
\end{proposition}

\begin{proofi} Let $\mathcal{D}$ be a  breadth-first $X$-chase derivation from $F$ and $\mathcal{R}$ and $\mathcal{D}_{|F'}$ the restriction of $\mathcal{D}$ induced by $F' \subseteq F$.

\noindent \textbf{Case O} Since $\mathcal{D}$ is breadth-first, it is rank compatible, and since the ordering of triggers is preserved in $\mathcal{D}_{|F'}$  we get that $\mathcal{D}_{|F'}$ is rank compatible. Similarly by the rank exhaustiveness of $\mathcal{D}$, all triggers which are descendants of $F'$ appear in $\mathcal{D}$, so $\mathcal{D}_{|F'}$ is also rank exhaustive. Hence $\mathcal{D}_{|F'}$ is breadth-first.

\noindent \textbf{Case SO} As in the \textbf{O} case, we can easily see that triggers in $\mathcal{D}_{|F'}$ are ordered by rank. Now,  suppose that $\mathcal{D}_{|F'}$ is not rank exhaustive, i.e., there are rule applications (descendants of $F'$) that were \textit{skipped} in $\mathcal{D}$ because they mapped the frontier variables of a rule $R$ in the same way that earlier rule applications \mlm{(using atoms from $F\setminus F'$)} did.
Then new triggers will be applicable in $\mathcal{D}_{|{F'}}$.

Let $\mathcal{D'}$ be a derivation, called the \textit{breadth first completion} of $\mathcal{D}_{|F'}$, constructed as follows:   for every breadth-first level $\kappa$, after  {sequentially} applying  all triggers of $\mathcal{D}_{|F'}$ of rank $\kappa$ that are still {\SO-applicable}, we complete this rank by applying all other possible $\mathbf{SO}$-applicable triggers of rank $\kappa$ (in any order). 

By construction, $\mathcal D'$ is a breadth-first \textbf{SO}-chase derivation. 
We will now show that it is actually a completion of  $\mathcal{D}_{|F'}$, in the sense that 
$\mathcal{D}_{|F'}$ is a subderivation of $\mathcal D'$. Indeed, suppose that the addition of a new trigger $(R,\pi)$ at rank $\kappa$ in $\mathcal{D'}$
cancels the \textbf{SO}-applicability of a trigger $(R,\pi')$ at rank  $\kappa' > \kappa$ in $\mathcal{D}_{|F'}$. So $(R,\pi)$ is ``frontier-equal'' with $(R,\pi')$.  Then, since $(R,\pi)$ is not in $\mathcal{D}$, and $\mathcal{D}$ is rank-exhaustive, there is a ``frontier-equal'' trigger $(R,\pi_D)$ in $\mathcal{D}$ at rank $\kappa_D \leq \kappa$; this is not possible since $(R,\pi_D)$ would also be frontier-equal to $(R,\pi')$, which would both belong to $\mathcal{D}$, which contradicts the fact that $\mathcal{D}$ is a \textbf{SO}-chase derivation. 


\medskip
\noindent \textbf{Case R} Let $\mathcal{D'}$ be the \textit{breadth first completion} of $\mathcal{D}_{|F'}$ constructed similarly as in the previous case:   for every breadth-first level $\kappa$, after sequentially applying all triggers of $\mathcal{D}_{|F'}$ of rank $\kappa$
that are still {\RE-applicable}, we complete this rank by applying all other possible $\RE$-applicable triggers of rank $\kappa$ (in any order). 
By construction, $\mathcal D'$ is a breadth-first \RE-chase derivation. 

\medskip
We will also show that $\mathcal{D}_{|F'}$  is a subderivation of $\mathcal{D'}$. 
%
We do so by contradiction. Let $(R,\pi)$ be the first trigger of $\mathcal{D}_{|F'}$ that does not appear in $\mathcal{D'}$. 

We denote by $\hat F'$ the resulting factbase after applying all the triggers that precede $(R,\pi)$ in $\mathcal{D}_{|F'}$ and by $G$ the resulting factbase after applying all triggers of $\mathcal{D}{'}$ up to $(R,\pi)$ {(excluding $(R,\pi)$)}. 
Let $(R_1,\pi_1),...,(R_m,\pi_m)$ be the triggers that were not $\mathbf{R}$-applicable in $\mathcal{D}$ but were $\mathbf{R}$-applicable in $\mathcal{D}'$ and added before $(R,\pi)$.
It holds that
$G=\hat F'\cup\pi^s_1(H_1)\cup\cdots\cup\pi_m^s(H_m)$.

\medskip Now, we have assumed that $(R,\pi)$ is not $\mathbf{R}$-applicable on $G$, hence not present in $\mathcal{D}'$. 
So, by the condition of $\mathbf{R}$-applicability, there exists a homomorphism 
$\sigma:\pi^s(H)\rightarrow G$ (so also $\sigma(\pi^s(H))\subseteq G$),
which behaves as the identity on $\pi(B)$.
We denote with $F_i$  the factbase produced just before applying $(R,\pi)$ on $\mathcal{D}$. 
We have that $\hat F'\subseteq F_i$, hence we get that 
$G\subseteq F_i \cup\pi^s_1(H_1)\cup\cdots\cup\pi_m^s(H_m)$ and therefore we also have
\begin{equation}\label{equ0}\sigma(\pi^s(H))\subseteq F_i \cup\pi^s_1(H_1)\cup\cdots\cup\pi_m^s(H_m)\end{equation}
Now, because  $(R_1,\pi_1),...,(R_m,\pi_m)$ were not $\mathbf{R}$-applicable in $\mathcal{D}$ we know that there exist respective homomorphisms $\sigma_j : \pi_j^s(H_j) \rightarrow   F_i$ (so also $\sigma_j ( \pi_j^s(H_j)) \subseteq  F_i$),  that behave as the identity on $\pi_j(B_j)$, for all $j\in\{1,...,m\}$.
As the domains of all $\sigma_j$ restricted to existential variables are disjoint, and $\sigma_j$ are the identity on non-existential variables, 
 we can define the substitution $\displaystyle\dot{\sigma}:=\bigcup_{i=1}^m\sigma_i$. By applying $\dot\sigma$ to both sides of (\ref{equ0}) we get
\begin{equation}\label{equ}\dot{\sigma}\circ\sigma(\pi^s(H))\subseteq \dot{\sigma}\big(F_i \cup\pi^s_1(H_1)\cup\cdots\cup\pi_m^s(H_m)\big)\end{equation} which, considering that $ \dot{\sigma}\big(F_i \cup\pi^s_1(H_1)\cup\cdots\cup\pi_m^s(H_m)\big) \subseteq F_i$, yields \begin{equation}\label{equ}\dot{\sigma}\circ\sigma(\pi^s(H))\subseteq F_i\end{equation} 
The homomorphism $\dot{\sigma}\circ\sigma$ can only substitute  
 the set of newly created variables in $\pi^s(H)$, hence qualifies as an extension of $\pi$, and from (\ref{equ}) we conclude that $(R,\pi)$ is not $\mathbf{R}$-applicable in $\mathcal{D}$. That is a contradiction, hence it must be the case that $(R,\pi)$ is indeed $\mathbf{R}$-applicable in $\mathcal{D}'$. Therefore we have shown that all triggers of  $\mathcal{D}_{|F'}$  appear in  $\mathcal{D}'$, so indeed  $\mathcal{D}_{|F'}$  is a subderivation of a breadth-first \RE-chase derivation from $F'$. 
\end{proofi}

\noindent 
The next property exploits the notion of consistent heredity to bound the size of the factbases that have to be considered. 

\begin{proposition}\label{prop-small-F}
\emph{Let $b$ be the maximum number of atoms in the bodies of the rules of a ruleset $\mathcal{R}$. Let X be any consistently hereditary chase. If there exist an $F$ and a  breadth-first X-chase $\mathcal{R}$-derivation from $F$ that is of depth at least $k$, then there exist an $F'$ of size $|F'|\leq b^{k}$ and a breadth-first X-chase $\mathcal{R}$-derivation from $F'$ with depth at least $k$. }
\end{proposition}


\begin{proofi} Let $\mathcal D$ be a breadth-first X-chase derivation from $F$ and $\mathcal{R}$ of depth $k$. Let $(R,\pi)$ be a trigger of $\mathcal{D}$ of depth $k$. Let $F'$ be the set of ancestors of $(R,\pi)$ in $F$, and by Lemma~\ref{anclue} we know that $|F'|\leq b^k$. Since the X-chase is consistently hereditary, the restriction $\mathcal D_{|F'}$ (which trivially includes $(R,\pi)$) is a subderivation of a breadth-first X-chase derivation $\mathcal{D}'$ from $F'$ and $\mathcal{R}$. According to the proof of proposition~\ref{conher}, for all three consistently hereditary chase variants, $\mathcal{D}'$ was constructed as a breadth-first completion of $\mathcal{D}$, therefore the ranks of common triggers are preserved from $\mathcal{D}$ to $\mathcal{D}_{|F'}$ and $\mathcal{D}'$. And since $\mathcal{D}'$ includes $(R,\pi)$ in its sequence of associated rule applications, we have that $(R,\pi)$ has also rank $k$ in $\mathcal{D}'$, hence $\mathcal{D}'$ is of depth at least $k$. 
\end{proofi}

\noindent We are now ready to prove the main result. 

\begin{theorem}\emph{Determining if a set of rules is X-$k$-bounded is decidable for any consistently hereditary chase variant X. This is in particular the case for the oblivious, semi-oblivious and restricted chase variants. }
\end{theorem}
\begin{proofi} By Proposition \ref{prop-small-F}, to check if all breadth-first X-chase derivations from $\mathcal R$ (with any factbase) are of depth at most $k$, it suffices to verify this property on all factbases of size less or equal to $b^{k}$. For a given factbase $F$, there is a finite number of (breadth-first) X-chase derivations from $F$ and $\mathcal R$ of depth at most $k$, hence we can effectively compute these derivations, and check if one of them can be extended to a derivation of depth $k+1$. 
\end{proofi}


\noindent Finally, the following example shows that the $\mathbf{E}$-chase (hence the core chase as well) is not consistently hereditary (hence not hereditary, as it the $\mathbf{E}$-chase is breadth-first).

\medskip
\noindent\textit{Example 11 (Equivalent chase).} 
Let $F=\{ s(b), p(a,a),  p(a,b), p(b,c)\}$ and $\mathcal{R}$ the following set of rules:   


\begin{tabular}{l}
	$R_1 = s(y) \wedge p(y,z) \wedge p(w,z) \wedge r(w) \rightarrow  q(w)$\\
	$R_2 = p(x,y) \wedge p(y,z) \rightarrow t(y)$\\
        $R_3 = p(x,x) \wedge p(x,y) \wedge p(y,z) \rightarrow\exists w \big(  p(w,z) \wedge r (w) \big) $\\
        $R_4 = t(y) \rightarrow  r(y)$\\
        $R_5 = p(x,y)  \rightarrow \exists u\ p(u,x)$

\end{tabular}

\noindent Here we can verify that any exhaustive E-chase derivation from $F$ and $\mathcal{R}$ is of depth 3.  Consider such a derivation $\mathcal D$ that adds atoms in the following specific order at each breadth-first level (for clarity, we do not use standardized names for the nulls):

\begin{tabular}{l}
	 0 : $s(b), p(a,a),  p(a,b), p(b,c)$\\
	 
          1 : $t(a), t(b), p(w_1,c), r(w_1), p(w_2,b), r(w_2), p(w_3,a), r(w_3)$\\

          2 : $q(w_1), r(a), r(b), p(u_1,w_1)$\\

          3 : $q(b)$\\
\end{tabular}

\noindent Below is a graphical representation of this derivation, where nodes are atoms and edges are colored according to different triggers:

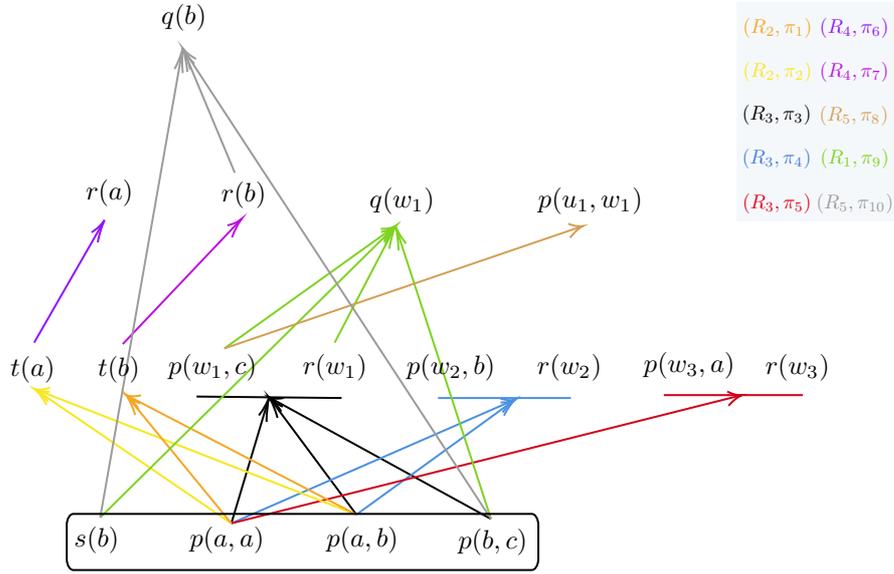
\tikzset{every picture/.style={line width=0.75pt}} 

\begin{tikzpicture}[x=0.75pt,y=0.75pt,yscale=-1.1,xscale=0.65]
\fill [color={rgb, 255:red, 245; green, 248; blue, 251 } ] (579,25) rectangle (705,125);
\draw    (164.67,205.33) -- (276,206) ;

\draw    (191.33,263.33) -- (219.43,207.45) ;
\draw [shift={(220.33,205.67)}, rotate = 476.7] [color={rgb, 255:red, 0; green, 0; blue, 0 }  ][line width=0.75]    (10.93,-3.29) .. controls (6.95,-1.4) and (3.31,-0.3) .. (0,0) .. controls (3.31,0.3) and (6.95,1.4) .. (10.93,3.29)   ;

\draw    (287.33,259.33) -- (221.89,206.92) ;
\draw [shift={(220.33,205.67)}, rotate = 398.69] [color={rgb, 255:red, 0; green, 0; blue, 0 }  ][line width=0.75]    (10.93,-3.29) .. controls (6.95,-1.4) and (3.31,-0.3) .. (0,0) .. controls (3.31,0.3) and (6.95,1.4) .. (10.93,3.29)   ;

\draw    (390.67,261.33) -- (222.23,206.29) ;
\draw [shift={(220.33,205.67)}, rotate = 378.1] [color={rgb, 255:red, 0; green, 0; blue, 0 }  ][line width=0.75]    (10.93,-3.29) .. controls (6.95,-1.4) and (3.31,-0.3) .. (0,0) .. controls (3.31,0.3) and (6.95,1.4) .. (10.93,3.29)   ;

\draw [color={rgb, 255:red, 74; green, 144; blue, 226 }  ,draw opacity=1 ]   (350,206) -- (452,206) ;

\draw [color={rgb, 255:red, 74; green, 144; blue, 226 }  ,draw opacity=1 ]   (191.33,263.33) -- (407.4,207.17) ;
\draw [shift={(409.33,206.67)}, rotate = 525.4300000000001] [color={rgb, 255:red, 74; green, 144; blue, 226 }  ,draw opacity=1 ][line width=0.75]    (10.93,-3.29) .. controls (6.95,-1.4) and (3.31,-0.3) .. (0,0) .. controls (3.31,0.3) and (6.95,1.4) .. (10.93,3.29)   ;

\draw [color={rgb, 255:red, 74; green, 144; blue, 226 }  ,draw opacity=1 ]   (287.33,259.33) -- (407.5,207.46) ;
\draw [shift={(409.33,206.67)}, rotate = 516.65] [color={rgb, 255:red, 74; green, 144; blue, 226 }  ,draw opacity=1 ][line width=0.75]    (10.93,-3.29) .. controls (6.95,-1.4) and (3.31,-0.3) .. (0,0) .. controls (3.31,0.3) and (6.95,1.4) .. (10.93,3.29)   ;

\draw [color={rgb, 255:red, 208; green, 2; blue, 27 }  ,draw opacity=1 ]   (191.33,263.33) -- (581.36,204.96) ;
\draw [shift={(583.33,204.67)}, rotate = 531.49] [color={rgb, 255:red, 208; green, 2; blue, 27 }  ,draw opacity=1 ][line width=0.75]    (10.93,-3.29) .. controls (6.95,-1.4) and (3.31,-0.3) .. (0,0) .. controls (3.31,0.3) and (6.95,1.4) .. (10.93,3.29)   ;

\draw [color={rgb, 255:red, 208; green, 2; blue, 27 }  ,draw opacity=1 ]   (523.33,204.67) -- (630,204.67) ;

\draw [color={rgb, 255:red, 248; green, 231; blue, 28 }  ,draw opacity=1 ]   (191.33,263.33) -- (42.52,202.75) ;
\draw [shift={(40.67,202)}, rotate = 382.15] [color={rgb, 255:red, 248; green, 231; blue, 28 }  ,draw opacity=1 ][line width=0.75]    (10.93,-3.29) .. controls (6.95,-1.4) and (3.31,-0.3) .. (0,0) .. controls (3.31,0.3) and (6.95,1.4) .. (10.93,3.29)   ;

\draw [color={rgb, 255:red, 248; green, 231; blue, 28 }  ,draw opacity=1 ]   (287.33,259.33) -- (42.61,202.45) ;
\draw [shift={(40.67,202)}, rotate = 373.09000000000003] [color={rgb, 255:red, 248; green, 231; blue, 28 }  ,draw opacity=1 ][line width=0.75]    (10.93,-3.29) .. controls (6.95,-1.4) and (3.31,-0.3) .. (0,0) .. controls (3.31,0.3) and (6.95,1.4) .. (10.93,3.29)   ;

\draw [color={rgb, 255:red, 245; green, 166; blue, 35 }  ,draw opacity=1 ]   (287.33,259.33) -- (112.58,205.26) ;
\draw [shift={(110.67,204.67)}, rotate = 377.19] [color={rgb, 255:red, 245; green, 166; blue, 35 }  ,draw opacity=1 ][line width=0.75]    (10.93,-3.29) .. controls (6.95,-1.4) and (3.31,-0.3) .. (0,0) .. controls (3.31,0.3) and (6.95,1.4) .. (10.93,3.29)   ;

\draw [color={rgb, 255:red, 245; green, 166; blue, 35 }  ,draw opacity=1 ]   (191.33,263.33) -- (112.28,205.84) ;
\draw [shift={(110.67,204.67)}, rotate = 396.03] [color={rgb, 255:red, 245; green, 166; blue, 35 }  ,draw opacity=1 ][line width=0.75]    (10.93,-3.29) .. controls (6.95,-1.4) and (3.31,-0.3) .. (0,0) .. controls (3.31,0.3) and (6.95,1.4) .. (10.93,3.29)   ;

\draw [color={rgb, 255:red, 144; green, 19; blue, 254 }  ,draw opacity=1 ]   (40,180.67) -- (92.61,126.11) ;
\draw [shift={(94,124.67)}, rotate = 493.96] [color={rgb, 255:red, 144; green, 19; blue, 254 }  ,draw opacity=1 ][line width=0.75]    (10.93,-3.29) .. controls (6.95,-1.4) and (3.31,-0.3) .. (0,0) .. controls (3.31,0.3) and (6.95,1.4) .. (10.93,3.29)   ;

\draw [color={rgb, 255:red, 189; green, 16; blue, 224 }  ,draw opacity=1 ]   (108,181.33) -- (197.64,125.06) ;
\draw [shift={(199.33,124)}, rotate = 507.88] [color={rgb, 255:red, 189; green, 16; blue, 224 }  ,draw opacity=1 ][line width=0.75]    (10.93,-3.29) .. controls (6.95,-1.4) and (3.31,-0.3) .. (0,0) .. controls (3.31,0.3) and (6.95,1.4) .. (10.93,3.29)   ;

\draw [color={rgb, 255:red, 126; green, 211; blue, 33 }  ,draw opacity=1 ]   (90.67,260.67) -- (315.61,128.35) ;
\draw [shift={(317.33,127.33)}, rotate = 509.53] [color={rgb, 255:red, 126; green, 211; blue, 33 }  ,draw opacity=1 ][line width=0.75]    (10.93,-3.29) .. controls (6.95,-1.4) and (3.31,-0.3) .. (0,0) .. controls (3.31,0.3) and (6.95,1.4) .. (10.93,3.29)   ;

\draw [color={rgb, 255:red, 126; green, 211; blue, 33 }  ,draw opacity=1 ]   (390.67,261.33) -- (318.29,129.09) ;
\draw [shift={(317.33,127.33)}, rotate = 421.31] [color={rgb, 255:red, 126; green, 211; blue, 33 }  ,draw opacity=1 ][line width=0.75]    (10.93,-3.29) .. controls (6.95,-1.4) and (3.31,-0.3) .. (0,0) .. controls (3.31,0.3) and (6.95,1.4) .. (10.93,3.29)   ;

\draw [color={rgb, 255:red, 126; green, 211; blue, 33 }  ,draw opacity=1 ]   (186,183.33) -- (315.49,128.12) ;
\draw [shift={(317.33,127.33)}, rotate = 516.9100000000001] [color={rgb, 255:red, 126; green, 211; blue, 33 }  ,draw opacity=1 ][line width=0.75]    (10.93,-3.29) .. controls (6.95,-1.4) and (3.31,-0.3) .. (0,0) .. controls (3.31,0.3) and (6.95,1.4) .. (10.93,3.29)   ;

\draw [color={rgb, 255:red, 126; green, 211; blue, 33 }  ,draw opacity=1 ]   (270.67,180.67) -- (316.02,128.84) ;
\draw [shift={(317.33,127.33)}, rotate = 491.19] [color={rgb, 255:red, 126; green, 211; blue, 33 }  ,draw opacity=1 ][line width=0.75]    (10.93,-3.29) .. controls (6.95,-1.4) and (3.31,-0.3) .. (0,0) .. controls (3.31,0.3) and (6.95,1.4) .. (10.93,3.29)   ;

\draw [color={rgb, 255:red, 212; green, 159; blue, 82 }  ,draw opacity=1 ]   (186,183.33) -- (459.37,127.73) ;
\draw [shift={(461.33,127.33)}, rotate = 528.5] [color={rgb, 255:red, 212; green, 159; blue, 82 }  ,draw opacity=1 ][line width=0.75]    (10.93,-3.29) .. controls (6.95,-1.4) and (3.31,-0.3) .. (0,0) .. controls (3.31,0.3) and (6.95,1.4) .. (10.93,3.29)   ;

\draw [color={rgb, 255:red, 155; green, 155; blue, 155 }  ,draw opacity=1 ]   (90.67,260.67) -- (153.43,47.92) ;
\draw [shift={(154,46)}, rotate = 466.44] [color={rgb, 255:red, 155; green, 155; blue, 155 }  ,draw opacity=1 ][line width=0.75]    (10.93,-3.29) .. controls (6.95,-1.4) and (3.31,-0.3) .. (0,0) .. controls (3.31,0.3) and (6.95,1.4) .. (10.93,3.29)   ;

\draw [color={rgb, 255:red, 155; green, 155; blue, 155 }  ,draw opacity=1 ]   (390.67,261.33) -- (155.48,47.35) ;
\draw [shift={(154,46)}, rotate = 402.3] [color={rgb, 255:red, 155; green, 155; blue, 155 }  ,draw opacity=1 ][line width=0.75]    (10.93,-3.29) .. controls (6.95,-1.4) and (3.31,-0.3) .. (0,0) .. controls (3.31,0.3) and (6.95,1.4) .. (10.93,3.29)   ;

\draw [color={rgb, 255:red, 155; green, 155; blue, 155 }  ,draw opacity=1 ]   (194,103) -- (155.15,47.64) ;
\draw [shift={(154,46)}, rotate = 414.94] [color={rgb, 255:red, 155; green, 155; blue, 155 }  ,draw opacity=1 ][line width=0.75]    (10.93,-3.29) .. controls (6.95,-1.4) and (3.31,-0.3) .. (0,0) .. controls (3.31,0.3) and (6.95,1.4) .. (10.93,3.29)   ;

\draw   (65,264.13) .. controls (65,261.3) and (67.3,259) .. (70.13,259) -- (421.87,259) .. controls (424.7,259) and (427,261.3) .. (427,264.13) -- (427,279.53) .. controls (427,282.37) and (424.7,284.67) .. (421.87,284.67) -- (70.13,284.67) .. controls (67.3,284.67) and (65,282.37) .. (65,279.53) -- cycle ;

\draw (292,271) node   {$p( a,b)$};
\draw (392,272) node   {$p( b,c)$};
\draw (188,271) node   {$p( a,a)$};
\draw (88,271) node   {$s( b)$};
\draw (98,112) node   {$r( a)$};
\draw (201,113) node   {$r( b)$};
\draw (322.33,116) node   {$q( w_{1})$};
\draw (467,116) node   {$p( u_{1} ,w_{1})$};
\draw (155,32) node   {$q( b)$};
\draw (39,193) node   {$t( a)$};
\draw (105,193) node   {$t( b)$};
\draw (177,192) node   {$p( w_{1} ,c)$};
\draw (271,192) node   {$r( w_{1})$};
\draw (451,192) node   {$r( w_{2})$};
\draw (360,192) node   {$p( w_{2} ,b)$};
\draw (543,191) node   {$p( w_{3} ,a)$};
\draw (626,192) node   {$r( w_{3})$};

\draw (610.62,36.71) node [scale=0.75,color={rgb, 255:red, 245; green, 166; blue, 35}  ,opacity=1 ]  {$( R_{2} ,\pi _{1})$};
\draw (610.62,56.71) node [scale=0.75,color={rgb, 255:red, 248; green, 231; blue, 28  } ,opacity=1 ]  {$( R_{2} ,\pi _{2})$};
\draw (610,76.57) node [scale=0.75,color=black  ,opacity=1 ]  {$( R_{3} ,\pi _{3})$};
\draw (610,96.57) node [scale=0.75,color={rgb, 255:red, 64; green, 124; blue, 220 }
  ,opacity=1 ]  {$( R_{3} ,\pi _{4})$};
\draw (610.19,116.86) node [scale=0.75,color={rgb, 255:red, 228; green, 2; blue, 27 }  ,opacity=1 ]  {$( R_{3} ,\pi _{5})$};
\draw (669.62,36.71) node [scale=0.75,color={rgb, 255:red, 144; green, 19; blue, 254 } ,opacity=1 ]  {$( R_{4} ,\pi _{6})$};
\draw (669.62,56.71) node [scale=0.75,color={rgb, 255:red, 189; green, 16; blue, 224 } ,opacity=1 ]  {$( R_{4} ,\pi _{7})$};
\draw (669.19,77) node [scale=0.75,color={rgb, 255:red, 212; green, 159; blue, 82 }  ,opacity=1 ]  {$( R_{5} ,\pi _{8})$};
\draw (669.9,96.71) node [scale=0.75,color={rgb, 255:red, 126; green, 211; blue, 33 }  ,opacity=1 ]  {$( R_{1} ,\pi _{9})$};
\draw (669.99,117) node [scale=0.75,color={rgb, 255:red, 155; green, 155; blue, 155 }  ,opacity=1 ]  {$( R_{5} ,\pi _{10})$};
\end{tikzpicture}


\noindent  At step 1, $R_2$ is applied twice, producing $t(a)$ and  $t(b)$, and $R_3$ is applied three times, producing  $p(w_1,c), r(w_1), p(w_2,b), r(w_2), p(w_3,a)$ and $r(w_3)$. Note that $R_1$ and $R_4$ are not applicable, and $R_5$ is not $\EQ$-applicable because it would produce redundant atoms. 
At step 2, $R_1$ is applied once (producing $q(w_1)$), $R_2$ and $R_3$ are not $\EQ$-applicable, $R_4$ is applied twice, and $R_5$ is applied once (producing $p(u_1,w_1)$). 
Finally, at step 3, $R_1$ is applied, which makes all further triggers redundant, hence no other rule is $\EQ$-applicable. 

\vspace{-15mm}

Let $F' = F \setminus \{s(b)\}$. Let $\mathcal D_{F'}$ be the restriction of $\mathcal D$ induced by $F'$.
Here is a graphical representation of $\mathcal D_{F'}$:

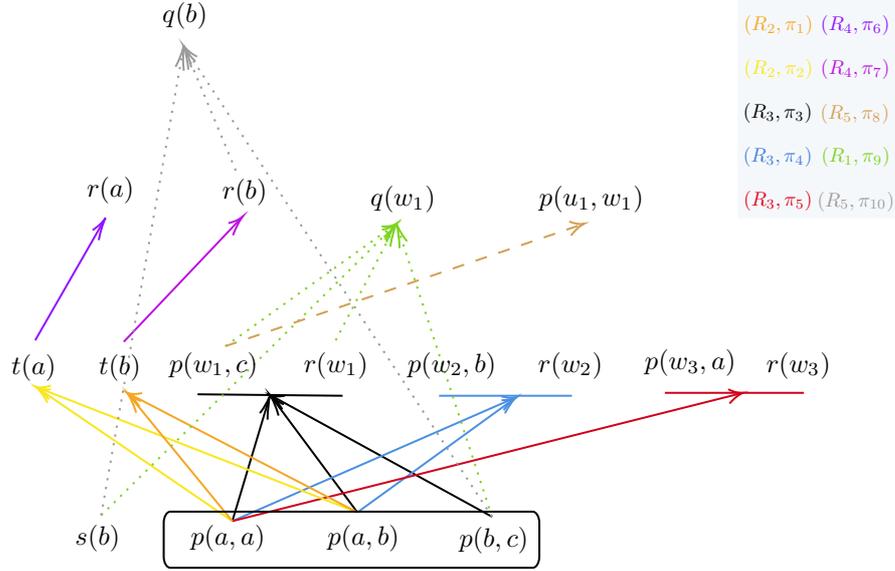
\tikzset{every picture/.style={line width=0.75pt}} 

\begin{tikzpicture}[x=0.75pt,y=0.75pt,yscale=-1.1,xscale=0.65]

\fill [color={rgb, 255:red, 245; green, 248; blue, 251 } ] (579,25) rectangle (705,125);
\draw    (164.67,205.33) -- (276,206) ;

\draw    (191.33,263.33) -- (219.43,207.45) ;
\draw [shift={(220.33,205.67)}, rotate = 476.7] [color={rgb, 255:red, 0; green, 0; blue, 0 }  ][line width=0.75]    (10.93,-3.29) .. controls (6.95,-1.4) and (3.31,-0.3) .. (0,0) .. controls (3.31,0.3) and (6.95,1.4) .. (10.93,3.29)   ;

\draw    (287.33,259.33) -- (221.89,206.92) ;
\draw [shift={(220.33,205.67)}, rotate = 398.69] [color={rgb, 255:red, 0; green, 0; blue, 0 }  ][line width=0.75]    (10.93,-3.29) .. controls (6.95,-1.4) and (3.31,-0.3) .. (0,0) .. controls (3.31,0.3) and (6.95,1.4) .. (10.93,3.29)   ;

\draw    (390.67,261.33) -- (222.23,206.29) ;
\draw [shift={(220.33,205.67)}, rotate = 378.1] [color={rgb, 255:red, 0; green, 0; blue, 0 }  ][line width=0.75]    (10.93,-3.29) .. controls (6.95,-1.4) and (3.31,-0.3) .. (0,0) .. controls (3.31,0.3) and (6.95,1.4) .. (10.93,3.29)   ;

\draw [color={rgb, 255:red, 74; green, 144; blue, 226 }  ,draw opacity=1 ]   (350,206) -- (452,206) ;

\draw [color={rgb, 255:red, 74; green, 144; blue, 226 }  ,draw opacity=1 ]   (191.33,263.33) -- (407.4,207.17) ;
\draw [shift={(409.33,206.67)}, rotate = 525.4300000000001] [color={rgb, 255:red, 74; green, 144; blue, 226 }  ,draw opacity=1 ][line width=0.75]    (10.93,-3.29) .. controls (6.95,-1.4) and (3.31,-0.3) .. (0,0) .. controls (3.31,0.3) and (6.95,1.4) .. (10.93,3.29)   ;

\draw [color={rgb, 255:red, 74; green, 144; blue, 226 }  ,draw opacity=1 ]   (287.33,259.33) -- (407.5,207.46) ;
\draw [shift={(409.33,206.67)}, rotate = 516.65] [color={rgb, 255:red, 74; green, 144; blue, 226 }  ,draw opacity=1 ][line width=0.75]    (10.93,-3.29) .. controls (6.95,-1.4) and (3.31,-0.3) .. (0,0) .. controls (3.31,0.3) and (6.95,1.4) .. (10.93,3.29)   ;

\draw [color={rgb, 255:red, 208; green, 2; blue, 27 }  ,draw opacity=1 ]   (191.33,263.33) -- (581.36,204.96) ;
\draw [shift={(583.33,204.67)}, rotate = 531.49] [color={rgb, 255:red, 208; green, 2; blue, 27 }  ,draw opacity=1 ][line width=0.75]    (10.93,-3.29) .. controls (6.95,-1.4) and (3.31,-0.3) .. (0,0) .. controls (3.31,0.3) and (6.95,1.4) .. (10.93,3.29)   ;

\draw [color={rgb, 255:red, 208; green, 2; blue, 27 }  ,draw opacity=1 ]   (523.33,204.67) -- (630,204.67) ;

\draw [color={rgb, 255:red, 248; green, 231; blue, 28 }  ,draw opacity=1 ]   (191.33,263.33) -- (42.52,202.75) ;
\draw [shift={(40.67,202)}, rotate = 382.15] [color={rgb, 255:red, 248; green, 231; blue, 28 }  ,draw opacity=1 ][line width=0.75]    (10.93,-3.29) .. controls (6.95,-1.4) and (3.31,-0.3) .. (0,0) .. controls (3.31,0.3) and (6.95,1.4) .. (10.93,3.29)   ;

\draw [color={rgb, 255:red, 248; green, 231; blue, 28 }  ,draw opacity=1 ]   (287.33,259.33) -- (42.61,202.45) ;
\draw [shift={(40.67,202)}, rotate = 373.09000000000003] [color={rgb, 255:red, 248; green, 231; blue, 28 }  ,draw opacity=1 ][line width=0.75]    (10.93,-3.29) .. controls (6.95,-1.4) and (3.31,-0.3) .. (0,0) .. controls (3.31,0.3) and (6.95,1.4) .. (10.93,3.29)   ;

\draw [color={rgb, 255:red, 245; green, 166; blue, 35 }  ,draw opacity=1 ]   (287.33,259.33) -- (112.58,205.26) ;
\draw [shift={(110.67,204.67)}, rotate = 377.19] [color={rgb, 255:red, 245; green, 166; blue, 35 }  ,draw opacity=1 ][line width=0.75]    (10.93,-3.29) .. controls (6.95,-1.4) and (3.31,-0.3) .. (0,0) .. controls (3.31,0.3) and (6.95,1.4) .. (10.93,3.29)   ;

\draw [color={rgb, 255:red, 245; green, 166; blue, 35 }  ,draw opacity=1 ]   (191.33,263.33) -- (112.28,205.84) ;
\draw [shift={(110.67,204.67)}, rotate = 396.03] [color={rgb, 255:red, 245; green, 166; blue, 35 }  ,draw opacity=1 ][line width=0.75]    (10.93,-3.29) .. controls (6.95,-1.4) and (3.31,-0.3) .. (0,0) .. controls (3.31,0.3) and (6.95,1.4) .. (10.93,3.29)   ;

\draw [color={rgb, 255:red, 144; green, 19; blue, 254 }  ,draw opacity=1 ]   (40,180.67) -- (92.61,126.11) ;
\draw [shift={(94,124.67)}, rotate = 493.96] [color={rgb, 255:red, 144; green, 19; blue, 254 }  ,draw opacity=1 ][line width=0.75]    (10.93,-3.29) .. controls (6.95,-1.4) and (3.31,-0.3) .. (0,0) .. controls (3.31,0.3) and (6.95,1.4) .. (10.93,3.29)   ;

\draw [color={rgb, 255:red, 189; green, 16; blue, 224 }  ,draw opacity=1 ]   (108,181.33) -- (197.64,125.06) ;
\draw [shift={(199.33,124)}, rotate = 507.88] [color={rgb, 255:red, 189; green, 16; blue, 224 }  ,draw opacity=1 ][line width=0.75]    (10.93,-3.29) .. controls (6.95,-1.4) and (3.31,-0.3) .. (0,0) .. controls (3.31,0.3) and (6.95,1.4) .. (10.93,3.29)   ;

\draw [color={rgb, 255:red, 126; green, 211; blue, 33 }  ,draw opacity=1 ] [dash pattern={on 0.84pt off 2.51pt}]  (90.67,260.67) -- (315.61,128.35) ;
\draw [shift={(317.33,127.33)}, rotate = 509.53] [color={rgb, 255:red, 126; green, 211; blue, 33 }  ,draw opacity=1 ][line width=0.75]    (10.93,-3.29) .. controls (6.95,-1.4) and (3.31,-0.3) .. (0,0) .. controls (3.31,0.3) and (6.95,1.4) .. (10.93,3.29)   ;

\draw [color={rgb, 255:red, 126; green, 211; blue, 33 }  ,draw opacity=1 ] [dash pattern={on 0.84pt off 2.51pt}]  (390.67,261.33) -- (318.29,129.09) ;
\draw [shift={(317.33,127.33)}, rotate = 421.31] [color={rgb, 255:red, 126; green, 211; blue, 33 }  ,draw opacity=1 ][line width=0.75]    (10.93,-3.29) .. controls (6.95,-1.4) and (3.31,-0.3) .. (0,0) .. controls (3.31,0.3) and (6.95,1.4) .. (10.93,3.29)   ;

\draw [color={rgb, 255:red, 126; green, 211; blue, 33 }  ,draw opacity=1 ] [dash pattern={on 0.84pt off 2.51pt}]  (186,183.33) -- (315.49,128.12) ;
\draw [shift={(317.33,127.33)}, rotate = 516.9100000000001] [color={rgb, 255:red, 126; green, 211; blue, 33 }  ,draw opacity=1 ][line width=0.75]    (10.93,-3.29) .. controls (6.95,-1.4) and (3.31,-0.3) .. (0,0) .. controls (3.31,0.3) and (6.95,1.4) .. (10.93,3.29)   ;

\draw [color={rgb, 255:red, 126; green, 211; blue, 33 }  ,draw opacity=1 ] [dash pattern={on 0.84pt off 2.51pt}]  (270.67,180.67) -- (316.02,128.84) ;
\draw [shift={(317.33,127.33)}, rotate = 491.19] [color={rgb, 255:red, 126; green, 211; blue, 33 }  ,draw opacity=1 ][line width=0.75]    (10.93,-3.29) .. controls (6.95,-1.4) and (3.31,-0.3) .. (0,0) .. controls (3.31,0.3) and (6.95,1.4) .. (10.93,3.29)   ;

\draw [color={rgb, 255:red, 212; green, 159; blue, 82 }  ,draw opacity=1 ] [dash pattern={on 4.5pt off 4.5pt}]  (186,183.33) -- (459.37,127.73) ;
\draw [shift={(461.33,127.33)}, rotate = 528.5] [color={rgb, 255:red, 212; green, 159; blue, 82 }  ,draw opacity=1 ][line width=0.75]    (10.93,-3.29) .. controls (6.95,-1.4) and (3.31,-0.3) .. (0,0) .. controls (3.31,0.3) and (6.95,1.4) .. (10.93,3.29)   ;

\draw [color={rgb, 255:red, 155; green, 155; blue, 155 }  ,draw opacity=1 ] [dash pattern={on 0.84pt off 2.51pt}]  (90.67,260.67) -- (153.43,47.92) ;
\draw [shift={(154,46)}, rotate = 466.44] [color={rgb, 255:red, 155; green, 155; blue, 155 }  ,draw opacity=1 ][line width=0.75]    (10.93,-3.29) .. controls (6.95,-1.4) and (3.31,-0.3) .. (0,0) .. controls (3.31,0.3) and (6.95,1.4) .. (10.93,3.29)   ;

\draw [color={rgb, 255:red, 155; green, 155; blue, 155 }  ,draw opacity=1 ] [dash pattern={on 0.84pt off 2.51pt}]  (390.67,261.33) -- (155.48,47.35) ;
\draw [shift={(154,46)}, rotate = 402.3] [color={rgb, 255:red, 155; green, 155; blue, 155 }  ,draw opacity=1 ][line width=0.75]    (10.93,-3.29) .. controls (6.95,-1.4) and (3.31,-0.3) .. (0,0) .. controls (3.31,0.3) and (6.95,1.4) .. (10.93,3.29)   ;

\draw [color={rgb, 255:red, 155; green, 155; blue, 155 }  ,draw opacity=1 ] [dash pattern={on 0.84pt off 2.51pt}]  (194,103) -- (155.15,47.64) ;
\draw [shift={(154,46)}, rotate = 414.94] [color={rgb, 255:red, 155; green, 155; blue, 155 }  ,draw opacity=1 ][line width=0.75]    (10.93,-3.29) .. controls (6.95,-1.4) and (3.31,-0.3) .. (0,0) .. controls (3.31,0.3) and (6.95,1.4) .. (10.93,3.29)   ;

\draw (292,271) node   {$p( a,b)$};
\draw (392,272) node   {$p( b,c)$};
\draw (188,271) node   {$p( a,a)$};
\draw (88,271) node   {$s( b)$};
\draw (98,112) node   {$r( a)$};
\draw (201,113) node   {$r( b)$};
\draw (322.33,116) node   {$q( w_{1})$};
\draw (467,116) node   {$p( u_{1} ,w_{1})$};
\draw (155,32) node   {$q( b)$};
\draw (39,193) node   {$t( a)$};
\draw (105,193) node   {$t( b)$};
\draw (177,192) node   {$p( w_{1} ,c)$};
\draw (271,192) node   {$r( w_{1})$};
\draw (451,192) node   {$r( w_{2})$};
\draw (360,192) node   {$p( w_{2} ,b)$};
\draw (543,191) node   {$p( w_{3} ,a)$};
\draw (626,192) node   {$r( w_{3})$};

\draw (610.62,36.71) node [scale=0.75,color={rgb, 255:red, 245; green, 166; blue, 35}  ,opacity=1 ]  {$( R_{2} ,\pi _{1})$};
\draw (610.62,56.71) node [scale=0.75,color={rgb, 255:red, 248; green, 231; blue, 28  } ,opacity=1 ]  {$( R_{2} ,\pi _{2})$};
\draw (610,76.57) node [scale=0.75,color=black  ,opacity=1 ]  {$( R_{3} ,\pi _{3})$};
\draw (610,96.57) node [scale=0.75,color={rgb, 255:red, 64; green, 124; blue, 220 }
  ,opacity=1 ]  {$( R_{3} ,\pi _{4})$};
\draw (610.19,116.86) node [scale=0.75,color={rgb, 255:red, 228; green, 2; blue, 27 }  ,opacity=1 ]  {$( R_{3} ,\pi _{5})$};
\draw (669.62,36.71) node [scale=0.75,color={rgb, 255:red, 144; green, 19; blue, 254 } ,opacity=1 ]  {$( R_{4} ,\pi _{6})$};
\draw (669.62,56.71) node [scale=0.75,color={rgb, 255:red, 189; green, 16; blue, 224 } ,opacity=1 ]  {$( R_{4} ,\pi _{7})$};
\draw (669.19,77) node [scale=0.75,color={rgb, 255:red, 212; green, 159; blue, 82 }  ,opacity=1 ]  {$( R_{5} ,\pi _{8})$};
\draw (669.9,96.71) node [scale=0.75,color={rgb, 255:red, 126; green, 211; blue, 33 }  ,opacity=1 ]  {$( R_{1} ,\pi _{9})$};
\draw (669.99,117) node [scale=0.75,color={rgb, 255:red, 155; green, 155; blue, 155 }  ,opacity=1 ]  {$( R_{5} ,\pi _{10})$};

\draw   (138.67,264.13) .. controls (138.67,261.3) and (140.96,259) .. (143.8,259) -- (421.87,259) .. controls (424.7,259) and (427,261.3) .. (427,264.13) -- (427,279.53) .. controls (427,282.37) and (424.7,284.67) .. (421.87,284.67) -- (143.8,284.67) .. controls (140.96,284.67) and (138.67,282.37) .. (138.67,279.53) -- cycle ;

\end{tikzpicture}

\noindent At level 2, $\mathcal D_{F'}$ still produces $r(a)$, $r(b)$ and $p(u_1,w_1)$ but not $q(w_1)$, and there is no step 3 because $R_1$ is not applicable. 
We can see that $\mathcal D_{F'}$ is not an \EQ-chase derivation because the application of $R_5$ at step 2 (which produces $p(u_1,w_1)$) is now redundant (this is due to the absence of $q(w_1)$). This already shows that the $\EQ$-chase is not hereditary. Moreover, we can check on $\mathcal D_{F'}$ that no rule application before the application of $R_5$ is able to add information on $w_1$ that would make $R_5$ $\EQ$-applicable at step 2.  Hence, $\mathcal D_{F'}$ is not contained in any $\EQ$-chase derivation from $F'$, which shows that the $\EQ$-chase is not consistently hereditary. 
Note also that any exhaustive \EQ-chase derivation from $F'$ is of depth 2 and not 3 as from $F$. 
\hfill $\triangle$\vspace{0.2cm}
%



%

\noindent\begin{minipage}{\linewidth}
\section{Conclusion}

In this paper, we investigated the problem of determining whether a
ruleset is $k$-bounded, that is when the chase always halts within a
predefined number of steps independently of the factbase. After discussing the concept of boundedness in breadth-first derivations, 
we have shown that $k$-boundedness is decidable for some important chase
variants by establishing a common property that ensures decidability,
namely ``consistent heredity''. The complexity of the problem is
independent from any data since the size of the factbases to be checked
depends only on $k$ and the size of the rule bodies.  
Our results indicate an EXPTIME upper bound for checking  $k$-boundedness for both the \textbf{O}-chase and the \textbf{SO}-chase.
For the \textbf{R}-chase, as the order of the rule applications matters,  one needs to check all possible derivations. This leads to a 2-EXPTIME upper bound for the \textbf{R}-chase.
We leave for further
work the study of the precise lower complexity bound  according to each kind of
chase. 
Finally, we leave open the question of the decidability of the $k$-boundedness for the core (or equivalent) chase.\end{minipage}
%
%
%
%
%
%
%
%
%

\bibliographystyle{alpha}
\bibliography{bibi}

\end{document}